\documentclass[11pt]{article}

\usepackage[preprint]{acl}
\usepackage{times}
\usepackage{latexsym}
\usepackage[T1]{fontenc}
\usepackage[utf8]{inputenc}
\usepackage{microtype}
\usepackage{inconsolata}
\usepackage{graphicx}

\usepackage[table]{xcolor}
\usepackage{array}
\usepackage{multirow}
 
\definecolor{headerblue}{RGB}{52, 90, 138}
\definecolor{rowgray}{RGB}{240, 240, 240}
\definecolor{avgrow}{RGB}{220, 230, 242}

\usepackage{mdframed}
\usepackage{kotex}
\usepackage{booktabs}
\usepackage{siunitx}  
\usepackage{comment} 
\usepackage{listings}
\usepackage{tabularx} 
\usepackage{xcolor} 
\usepackage{enumitem}
\usepackage{multirow}

\usepackage{microtype}
\usepackage{graphicx}
\usepackage{subcaption}
\usepackage{hyperref}
\usepackage{enumitem}

\usepackage{amsmath}
\usepackage{amssymb}
\usepackage{amsthm}
\usepackage[capitalize,noabbrev]{cleveref}
\theoremstyle{plain}

\theoremstyle{definition}

\theoremstyle{remark}

\usepackage[textsize=tiny]{todonotes}
\usepackage{makecell}
\usepackage{comment}
\usepackage{fvextra}
\usepackage{float}
\usepackage[most]{tcolorbox}   

\DefineVerbatimEnvironment{PromptBox}{Verbatim}{
  breaklines=true,
  breakanywhere=true,
  fontsize=\small,
  frame=single,
  framesep=4pt
}

\newcommand{\promptfile}[2]{%
  \begin{tcolorbox}[
      breakable, enhanced,
      title={#1}, fonttitle=\small\bfseries,
      colback=gray!3, colframe=gray!60,
      left=2mm, right=2mm, top=1mm, bottom=1mm]
  \VerbatimInput[
      breaklines, breakanywhere,
      fontsize=\scriptsize,
      tabsize=4]{latex/prompts/tex/#2.txt}%
  \end{tcolorbox}%
}

\newcolumntype{L}[1]{>{\raggedright\arraybackslash}p{#1}}

\title{Evolve Together, Fall into Place: How Social Scaffolding Shapes Self-Evolving Multi-Agent Systems}

\author{
  Seojin Hwang \\
  Chung-Ang University \\
  \texttt{swiftie1230@cau.ac.kr} \\\And
  Youngmin Choi \\
  Chung-Ang University \\
  \texttt{martinus.choi@gmail.com} \\\And
  Hwanhee Lee\thanks{Corresponding author.} \\
  Chung-Ang University \\
  \texttt{hwanheelee@cau.ac.kr}
}

\begin{document}
\maketitle

\begin{abstract}
Recent work lets LLM agents evolve their own harness, up to the team itself.
Yet in human groups, the same abilities achieve different outcomes under different social conditions.
Borrowing the notion of scaffolding from developmental psychology, we investigate whether the \emph{social scaffolding} of a self-evolving multi-agent system (MAS), namely its \emph{team composition} and \emph{structure of interdependence} (e.g., collaboration, conformity, or competition), shapes how the team evolves.
We present \textbf{AGORA} (\textbf{A}gents \textbf{G}rowing through \textbf{O}rganization and \textbf{R}elational \textbf{A}rrangements), a framework that varies the two axes independently across three evolution mechanisms.
Across web search, codebase understanding, and document QA, we find that (1) evolution alone rarely helps, while the right scaffolding surpasses a strong agent evolved alone by up to 25 points; (2) it does so by moving authority to the most capable member while the others step back; and (3) this member, having grown within the team, outperforms the same model alone. Thus, designing a self-evolving MAS requires designing its social scaffolding, which shapes not only how its agents learn but also who gets to lead.
\end{abstract}
\section{Introduction}

As Large Language Model (LLM)  agents take on increasingly complex tasks, they rarely work as bare models. They run inside a harness that shapes their performance~\cite{meng2026agent}.
Prior work has focused on its technical design, from what a single agent sees and is allowed to do~\cite{lee2026meta,lou2026autoharness} to how LLM calls are composed into workflows~\cite{zhang2025aflow}. The same holds for multi-agent systems (MAS)~\cite{liu2026synthesizing}, and a growing line of work goes further and lets agents evolve their own harness, by rewriting their skills, memories, and even the team itself~\cite{hao2026evolve,yang2026skillopt,zhang2026memrl}.

\begin{figure}[!ht]
\centering
\includegraphics[width=\linewidth]{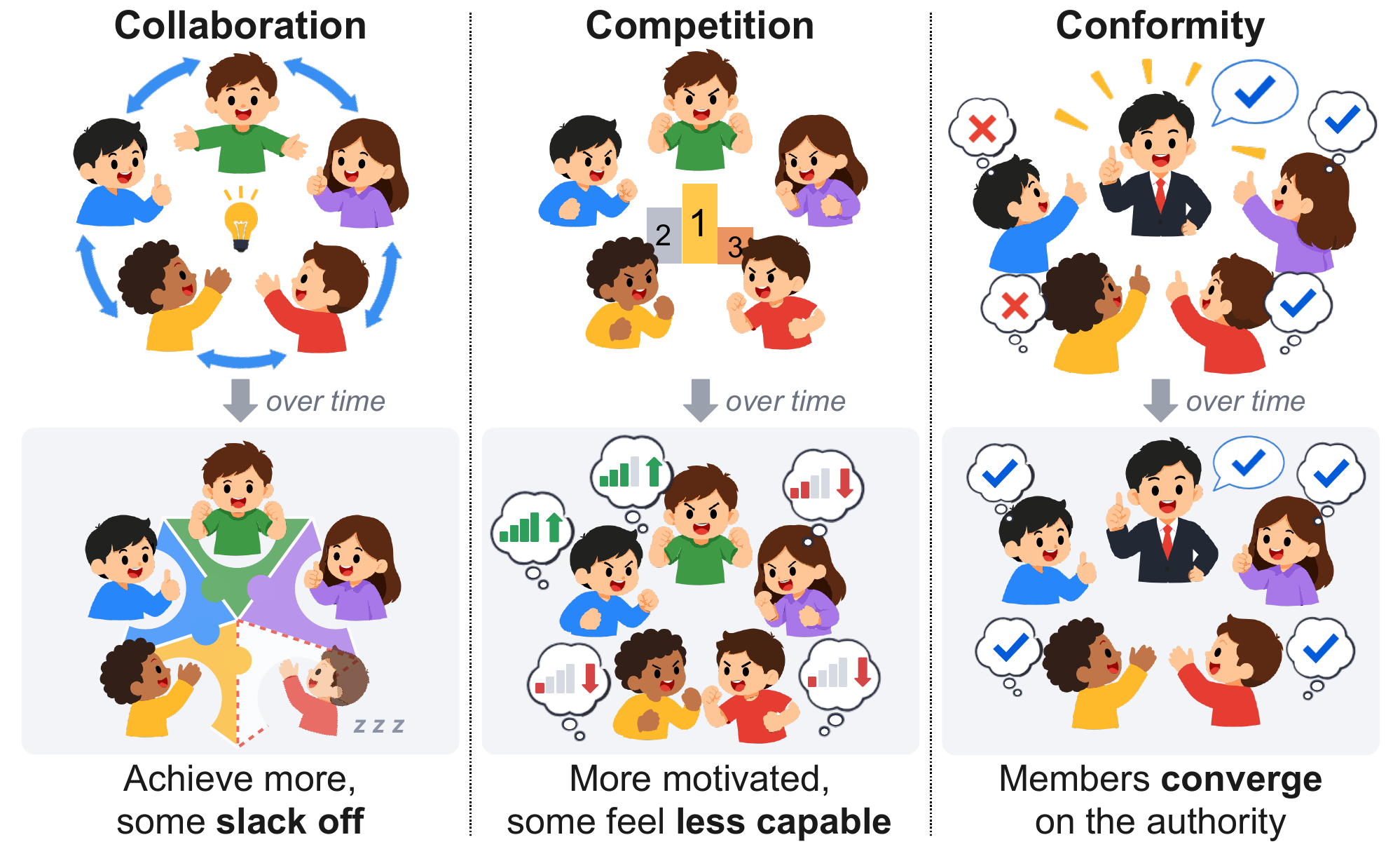}
\vspace{-5mm}
\caption{How different social scaffolding leads to different evolutionary outcomes in human groups.
}

\vspace{-5mm}
\label{fig:intro-challenge}
\end{figure}

However, as shown in Figure~\ref{fig:intro-challenge}, evolution in human groups never happens in a social vacuum.
Organizational and evolutionary theories suggest that people with the same abilities can achieve and behave differently, depending on how they compare with those around them and how much they rely on one another~\cite{deutsch1949theory,johnson2009educational}.
Much like a human group, a self-evolving MAS consists of agents that differ in ability, depend on one another, and change through experience. This raises a natural question: \textit{does the social layer of its harness shape how the team evolves, as social conditions do for human groups?}

Prior work on self-evolving MAS either fixes this social layer as a design~\cite{liu2026spiralselfplayzerosumgames, chen2025multiagentevolvellmselfimprove, pan2025advevomarlshapinginternalizedsafety} or folds it into the evolution process itself~\cite{hao2026evolve}. We instead treat this social layer as the central variable of study. We borrow the developmental-psychology notion of scaffolding, the support a learner's social environment provides during development and withdraws once it is no longer needed~\cite{wood1976role}, and call this layer \textbf{\emph{social scaffolding}}. We operationalize it for MAS along the two factors above: \emph{team composition}, which sets how capable each agent is relative to its teammates, and the \emph{structure of interdependence}, which sets how agents share information and decide on team changes.

To isolate its effect on what a team evolves into, we build \textbf{AGORA} (\textbf{A}gents \textbf{G}rowing through \textbf{O}rganization and \textbf{R}elational \textbf{A}rrangements), a framework that extends the three-level evolution loop of Meta-Team~\citep{hao2026evolve}. After each batch of training tasks, every agent improves itself (L1), agents take notes on their teammates when a structure of interdependence is in place (L2), and the team revises itself as a whole (L3), with each team revision checked on a held-out set.

AGORA varies the two axes independently: team composition sets the mix of agent capabilities (onlyweak, onlystrong, or weakstrong), and the structure of interdependence (none, collaboration, conformity, or competition) sets which permissions agents hold, how they share information at L2, and how team changes are decided at L3. We cross these axes with three evolution mechanisms that govern L1, Meta-Team~\cite{hao2026evolve}, SkillOpt~\cite{yang2026skillopt}, and MemRL~\cite{zhang2026memrl}, holding everything else fixed so that differences can be traced to social scaffolding.

Using AGORA on three benchmarks spanning web search (GAIA), codebase understanding (LoCoBench), and document QA (OfficeQA Pro), we investigate three research questions: whether social scaffolding changes what a self-evolving MAS achieves (RQ1); how the team reorganizes under the right scaffolding (RQ2); and why the reorganized team outperforms the same model working alone (RQ3). Experiments yield three findings.
\begin{itemize}[leftmargin=*, topsep=0pt, itemsep=2pt, parsep=0pt]
    \item \textbf{Evolution alone is rarely enough (RQ1).} Evolving under no structure of interdependence rarely improves a team, whereas the right social scaffolding does, surpassing a strong agent evolved alone by up to 25.0 points. Which structure helps depends on who is in the team.
    \item \textbf{Authority moves to the most capable member (RQ2).} Effective scaffolding hands the chair to the most capable member, as human groups gain from recognizing expertise, and the others step back unless they remove a bottleneck. Unlike human groups, they drop out rather than loaf.
    \item \textbf{The team grows its leader (RQ3).} The member left leading, having learned from its teammates, outperforms the same model evolved alone.
\end{itemize}
These findings show that plugging an evolution mechanism into an off-the-shelf MAS is not enough; the social scaffolding under which the team evolves matters just as much, deciding not only how its agents learn, but who gets to lead.

\section{Related Work}
\begin{figure*}[ht]
  \centering
  \includegraphics[width=\textwidth]{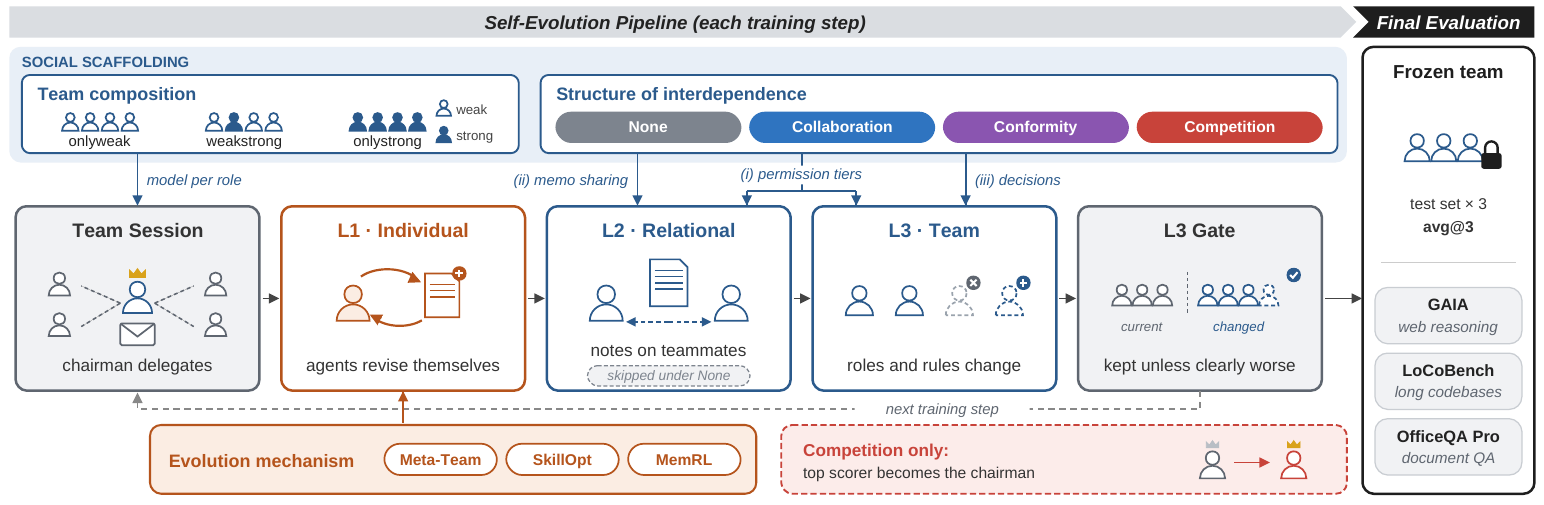}
  \vspace{-5mm}
  \caption{Overview of AGORA. After completing a task as a team (\textit{Team Session}), each agent updates itself (\textit{L1}) and writes memos about its peers, guided by social scaffolding (\textit{L2}). Based on these memos, team-level structural changes are proposed (\textit{L3}) and then retained or rejected by a held-out gate (\textit{L3 Gate}). Throughout this process, we vary the social scaffolding and the evolution mechanism to examine how a self-evolving MAS evolves.}
  
  \label{fig:AGORA}
  \vspace{-4mm}
\end{figure*}

\subsection{Social Dynamics in Human Groups}
In human groups, individuals with identical abilities can reach different outcomes depending on their social scaffolding.
\textit{Collaboration} raises achievement over competing or working alone~\cite{deutsch1949theory,johnson2009educational}, but members may loaf when the work is shared~\cite{latane1979many,karau1993social}.
\textit{Competition} raises motivation, but equally able members feel less capable among stronger peers~\cite{marsh1987big}.
\textit{Conformity} reduces individual error but can suppress innovation~\cite{henrich1998evolution}, and since people obey even wrong authority~\cite{milgram1963behavioral} and groups that suppress dissent decide poorly~\cite{janis1972victims}, the outcome depends on the decision-maker.

\subsection{Evolution and Social Dynamics in LLM Agents}
Evolving agents improve their external state without updating model weights, refining memory~\cite{zhu2026unified,choi2026membrane,zhang2026memrl,wang2026sage,liu2026evolvemem} or skills~\cite{zhang2026coevoskills,zhu2026esc,zhang2026evoagent,yang2026skillopt}, and recently agent composition and interaction~\cite{yuan2025evoagent,lee2025evolving,dang2026multi}. MAS gain more over single agents when their sub-agents are weaker~\cite{ke2026mas}, suggesting that well-organized weak agents can close much of the gap to stronger ones.

In MAS, collaboration does not always help~\cite{tran2026singleagentllmsoutperformmultiagent,pappu2026multiagentteamsholdexperts}, competition can increase systemic failure~\cite{mori2026aiagentswalkbar}, and conformity varies with group structure and roles~\cite{weng2025we,cho2025herd,choi2025empiricalstudygroupconformity,bito2026largelanguagemodelsexhibit,han2026conformitydynamicsllmmultiagent,bellina2026conformitysocialimpactai,choi-etal-2026-belief}. Social structures are also used to improve agents through self-play, mutual checks~\cite{liu2026spiralselfplayzerosumgames,chen2025multiagentevolvellmselfimprove,pan2025advevomarlshapinginternalizedsafety}, or orchestration~\cite{qi2026individualintelligencesurveyingcollaboration}, and Meta-Team~\cite{hao2026evolve} jointly evolves the agents, their interactions, and the team. These methods, however, fix a single interdependence structure, often on closed-answer tasks. AGORA instead turns this social layer into a controlled variable, varying team composition and the structure of interdependence on open-ended, real-world tasks.
\section{AGORA Framework}

\begin{table*}[t]
\centering
\small
\resizebox{\textwidth}{!}{
\begin{tabular}{lp{3.6cm}p{3.6cm}p{3.4cm}p{1.5cm}}
\toprule
Level & What changes & Who changes it & Takes effect & Gate \\
\midrule
\multirow{3}{*}{L1 (individual)}
    & Skill document & Mechanism (SkillOpt) & Next iteration & L1 gate \\
\cmidrule(l){2-5}
    & Memory bank & Mechanism (MemRL) & Next iteration & None \\
\cmidrule(l){2-5}
    & Prompt patch & Agent itself (Meta-Team) & Next task & None \\
\midrule
L2 (relational) & L2 memo & Each agent, about teammates it worked with & Next iteration & None \\
\midrule
L3 (team) & Roles, agent prompts and configurations, constitution & Proposed by Tier~1 agents & Next iteration, unless reverted by the L3 gate & L3 gate \\
\bottomrule
\end{tabular}
}
\caption{Levels of the AGORA operation. L1 and L3 evolve the agents and the team, while L2 records relations between agents only when the social scaffolding includes interdependence. Social scaffolding acts directly only on L2 and L3, and reaches L1 through the trajectories and L2 memos that every mechanism learns from.}
\vspace{-5mm}
\label{tab:layers}
\end{table*}

As shown in Figure~\ref{fig:AGORA}, to test whether social scaffolding shapes how a self-evolving MAS evolves, we design AGORA as a controlled testbed. AGORA keeps the basic loop of Meta-Team~\citep{hao2026evolve}, in which a chairman-led team evolves its external state at the individual (L1), relational (L2), and team (L3) levels, but turns its fixed social layer, where the chairman alone decides team changes, into the variable under study. 
AGORA varies team composition and the structure of interdependence while swapping in different L1 evolution mechanisms, separating their effects.

\subsection{Overall Framework Formulation}

\subsubsection{Team Definition}
A team is a set of role-specific agents, defined by a team configuration (e.g., chairman, team budgets), a constitution of team-wide rules that can be revised as the team evolves, and the agents themselves. Each agent has a configuration (e.g., model and tools), a prompt, and an evolution state such as a skill document or a memory bank. The chairman receives the task, calls on the other agents, and submits the final answer.

\subsubsection{Session Execution}
Each task is handled in a single team session, whose final answer from the chairman is scored. Within a session, agents operate concurrently rather than taking turns, interleaving tool calls (e.g., web search) with messages through a shared mailbox, under a budget on messages, and time that is reported after every call. Detailed limits are given in Appendices~\ref{app:team_config} and~\ref{app:engine}.

\subsubsection{Three-Level Operation}
Under a given social scaffolding, AGORA repeats an evolution loop of a team session followed by three levels of evolution and a held-out gate (Figure~\ref{fig:AGORA} and Table~\ref{tab:layers}). L1 improves each agent through its own evolution mechanism, L2 records what agents learn about one another, L3 revises the team as a whole, and the L3 gate reverts a revision only if it clearly hurts the team on held-out data.


\paragraph{L1 evolution (Individual)} After completing a task as a team (Figure~\ref{fig:AGORA}, \textit{`Team Session'}), each agent evolves through its own evolution mechanism (Figure~\ref{fig:AGORA}, \textit{`L1 $\cdot$ Individual'}). Specifically, it updates a skill with its gate record in SkillOpt, a memory bank of Q-valued entries in MemRL, and a prompt patch with self-written skills in Meta-Team. Unlike L2 and L3, L1 operates in the same way under every social scaffolding; the scaffolding affects it only indirectly, through the trajectories and L2 memos from which the agents learn.


\paragraph{L2 recording (Relational)}
L2 models what teammates learn about one another through social interaction (Figure~\ref{fig:AGORA}, \textit{`L2 $\cdot$ Relational'}). Once L1 has finished, each agent writes a memo on every teammate it has worked with, recording strengths, weaknesses, collaboration notes, and contradictions; a judge model flags contradictions but does not block the memo. How much agents share and whether memos are reviewed before saving depend on the \textit{structure of interdependence} (Section~\ref{sec:interdependence}), and L2 is skipped under \textit{none}. Memos enter the author's system prompt in later sessions, inform L3 revisions, and are passed to the update models of SkillOpt and MemRL.


\paragraph{L3 evolution (Team)} After each agent has written memos on its teammates in L2, L3 evolves the team as a whole based on these memos, by adding or removing roles and revising the agents' prompts and configurations, and the constitution. Tier~1 agents propose revisions, which are aggregated according to the social scaffolding and kept unless the L3 gate judges the revised team to be clearly worse (Figure~\ref{fig:AGORA}, \textit{`L3 Gate'}). Details of Tier~1 are provided in Section~\ref{sec:interdependence}.


\begin{table*}[t]
\centering
\small
\begin{tabular}{p{0.15\textwidth}p{0.25\textwidth}p{0.25\textwidth}p{0.25\textwidth}}
\toprule
 & Meta-Team & SkillOpt & MemRL \\
\midrule
What evolves & Prompt patches and skills written by the agent & Skill document & Memory bank with Q-values \\
\midrule
Injected into execution & Own patches and skill list & Current skill document & Up to 5 retrieved memories (success/failure scripts) \\
\midrule
Evolution timing & Right after each task & End of iteration & End of iteration \\
\midrule
Evolution signal & Agent's self-reflection & Reflection on trajectories with grading results & Task success or failure as reward \\
\midrule
L1 gate & None & Held-out binomial test & None \\
\midrule
Model performing the update & Agent's own model (weak or strong) & The strong model (fixed), with L2 notes & The strong model (fixed), with L2 notes \\
\midrule
Steps $\times$ batch & $1 \times (17 / 20)$ & $4 \times (17 / 20)$ & $4 \times (17 / 20)$ \\
\midrule
Batch execution & Sequential & Concurrent & Concurrent \\
\bottomrule
\end{tabular}
\caption{Evolution mechanisms. Batch sizes are 17 for GAIA and 20 for LoCoBench and OfficeQA Pro. All mechanisms share the same L2 and L3 and the same L3 gate.}
\vspace{-3mm}
\label{tab:mechanisms}
\end{table*}

\subsubsection{L3 Held-out Gate}
\label{sec:gate}
The L3 held-out gate reverts a team revision only if it clearly harms the team. The current and revised teams each solve the same 20 validation problems with the same seeds, so that the two are compared on equal footing (Figure~\ref{fig:AGORA}, \textit{`L3 Gate'}). The validation set is disjoint from the train and test sets, and the gate is applied identically under every evolution mechanism and structure of interdependence.

All L3 changes of an iteration are reverted if the revised team succeeds less often and a one-sided binomial test against the current team's success rate finds the drop significant. For example, if the current team solves 10 of the 20 problems, the changes are reverted only if the revised team solves 8 or fewer.
This is because L3 revisions are decisions the team made under its own structure of interdependence; it blocks only clear harm and leaves the rest to the team. We use a relaxed $\alpha = 0.3$ because re-running the same team changes the outcome of about a third of the validation problems, so a strict level would rarely catch even real degradation. Details about thresholds are in Appendix~\ref{app:gate}. Each iteration also applies at most two role additions or deletions and two patch bundles, so that each gate judges only a few changes and the team cannot grow without bound.

\subsection{Evolution Mechanism}\label{sec:mechanism}
As shown in Table~\ref{tab:mechanisms}, we include three L1 mechanisms that differ in what they update and what drives the update: an agent's own reflection (Meta-Team), reflection on graded trajectories (SkillOpt), and task reward on a memory bank (MemRL). We adapt the original implementations to the multi-agent setting with minimal changes. The mechanism determines only L1. 
Details and all changes from the original are in Appendix~\ref{app:evolution}.

\paragraph{Meta-Team}
Following the original design, Meta-Team runs a single iteration with its batch executed sequentially. Right after each task, each agent reflects on its own behavior and edits its own prompt patch and skills with its own model, and the edits take effect immediately without a gate.

\paragraph{SkillOpt}
SkillOpt runs for four iterations. At the end of each iteration, each role applies reflect, merge, and apply steps to its trajectories from that iteration, both successful and failed, together with their grading results, to produce a candidate skill document. The candidate replaces the current document in the role's system prompt only if it passes the L1 gate, that is, only if it clearly improves performance on the validation set (Appendix~\ref{app:gate}).

\paragraph{MemRL}
MemRL runs for four iterations, with a separate memory bank for each role. Before a session, up to five memories are retrieved by combining their similarity to the problem with their Q-values. At the end of each iteration, the Q-values of the retrieved memories are updated according to task success, and each trajectory, successful or not, is summarized as a procedural script and added as a new memory. Updates take effect without a gate, so the bank keeps both what worked and what failed.

\subsection{Social Scaffolding Implementation}

In human teams, how members grow depends both on the capabilities of their teammates, including the gaps between them, and on how they depend on one another while working together. Following this, we implement social scaffolding along two axes: team composition and the structure of interdependence (Figure~\ref{fig:AGORA}, top).

\begin{table*}[t]
\centering
\small
\begin{tabular*}{\textwidth}{@{\extracolsep{\fill}}llllll@{}}
\toprule
Intervention & Aspect & None & Collaboration & Conformity & Competition \\
\midrule
\multirow{3}{*}{Permission tiers}
    & Chairman & Fixed & Fixed & Fixed & Rotates to top scorer \\
\cmidrule(l){2-6}
    & Tier~1 roles & All roles & All roles & Chairman only & Chairman only \\
\midrule
\multirow{2}{*}{L2 memo}
    & Sharing & Disabled & Active & When stuck; deferential & Minimal; scores shown \\
\cmidrule(l){2-6}
    & Finalization & Disabled & After peer review & Immediately & After peer review \\
\midrule
\multirow{2}{*}{L3 team config}
    & Proposals & All roles & All roles & Chairman only & Chairman only \\
\cmidrule(l){2-6}
    & Peer review & No & Majority & No & Majority \\
\cmidrule(l){2-6}
    & Decision rule & Keep all & Review tally & Chairman alone & Chairman, given reviews \\
\bottomrule
\end{tabular*}
\vspace{-1mm}
\caption{Structures of interdependence. `Majority' peer review passes once a majority of participating teammates have responded, and Tier~2 roles can only suggest L3 changes to the chairman. Details are given in Section~\ref{sec:interdependence}.}
\vspace{-5mm}
\label{tab:social_main}
\end{table*}

\subsubsection{Team Composition}\label{sec:composition}
To realize capability differences among team members, we use two model tiers, weak and strong, drawn from the same model family so that the capability gap remains the main variable and confounds from mixing model families are avoided. As shown in '\textit{Team composition}' in Figure~\ref{fig:AGORA}, we compare three configurations: \textbf{onlyweak}, \textbf{onlystrong}, and \textbf{weakstrong}. Onlyweak and onlystrong assign the weak and strong model, respectively, to every role. Weakstrong upgrades only the bottleneck role of each benchmark to the strong model (web\_agent for GAIA, developer for LoCoBench, and reader for OfficeQA), while all other roles, including the chairman, remain weak. 

On GAIA, varying which and how many roles are strong shows that which role matters far more than how many: before evolution, a strong bottleneck role alone gives 48.3, a strong non-bottleneck role at most 28.3, and further strong roles barely help (51.7--52.5). Details on model selection, bottleneck roles, role tools, and these variants are in Appendices~\ref{app:team_config} and~\ref{app:res_comp}, and representative role prompts and the full constitution in Appendix~\ref{app:prompts}.

\subsubsection{Structure of Interdependence}\label{sec:interdependence}
To realize how members depend on one another, we define four conditions: \textbf{none}, \textbf{collaboration}, \textbf{conformity}, and \textbf{competition}. As shown in Table~\ref{tab:social_main}, each condition differs not only in prompt wording but in tools, permissions, and decision rules, and intervenes at three corresponding points of the loop, marked (i)--(iii) in Figure~\ref{fig:AGORA}. The three points are (i) \emph{permission tiers}, which set who is Tier~1 and can read the whole team's memos, while Tier~2 agents read only their own; (ii) \emph{L2 memos}, which set how much agents share and whether memos are saved directly or staged for objection before finalized; and (iii) \emph{L3 decisions}, which set who can propose, whether proposals go through peer review, and who makes the end-of-iteration decision by what rule.

\paragraph{None.}
A baseline without social interaction: all agents are Tier~1, there is no L2 (skipped under \textit{None} in Figure~\ref{fig:AGORA}), and all L3 proposals are kept at the end-of-iteration decision.

\paragraph{Collaboration.}
All agents are Tier~1 and share actively, and both memos and L3 proposals go through peer review. A memo is finalized once every teammate the author interacted with in the task has responded, and an L3 proposal passes once a majority of the participating teammates have commented on it. The end-of-iteration decision tallies the SUPPORT and OPPOSE reviews.

\paragraph{Conformity.}
Only the chairman is Tier~1 and decides alone, and agents have no tool for raising objections. In L2, agents defer to their teammates' self-assessments, contact teammates only when stuck, and save memos directly.

\paragraph{Competition.}
Agents share minimally and are shown comparative scores of their own contribution. Memos and L3 proposals go through the same peer review as in collaboration, with objections citing these scores, but the chairman decides alone at the end of each iteration. The role with the sole top score then becomes Tier~1 and the chairman for the next iteration, as shown by \textit{`Competition only'} in Figure~\ref{fig:AGORA}. This rotation is applied after the L3 gate, outside L3, and is not itself gated.

\providecommand{\tbd}{\textcolor{gray}{\textit{tbd}}}
\begin{table*}[t]
\centering
\setlength{\tabcolsep}{5pt}
\resizebox{\textwidth}{!}{%
\begin{tabular}{ll c|ccc c|ccc c|ccc}
\toprule
& & \multicolumn{4}{c}{GAIA} & \multicolumn{4}{c}{LoCoBench} & \multicolumn{4}{c}{OfficeQA Pro} \\
\cmidrule(lr){3-6} \cmidrule(lr){7-10} \cmidrule(lr){11-14}
Team & Structure & Pre & Meta-Team & SkillOpt & MemRL & Pre & Meta-Team & SkillOpt & MemRL & Pre & Meta-Team & SkillOpt & MemRL \\
\midrule
solo\_weak & None & 20.8\,{\scriptsize$\pm$1.4} & \cellcolor[RGB]{251,246,236}20.8\,{\scriptsize$\pm$5.2} & \cellcolor[RGB]{251,246,236}20.8\,{\scriptsize$\pm$5.2} & \cellcolor[RGB]{243,240,233}20.0\,{\scriptsize$\pm$0.0} & 54.5\,{\scriptsize$\pm$2.0} & \cellcolor[RGB]{212,219,224}50.8\,{\scriptsize$\pm$1.2} & \cellcolor[RGB]{198,210,219}49.5\,{\scriptsize$\pm$3.6} & \cellcolor[RGB]{170,191,210}46.8\,{\scriptsize$\pm$3.8} & 2.7\,{\scriptsize$\pm$0.9} & \cellcolor[RGB]{251,246,236}2.7\,{\scriptsize$\pm$0.9} & \cellcolor[RGB]{241,214,208}5.9\,{\scriptsize$\pm$0.9} & \cellcolor[RGB]{246,242,234}2.2\,{\scriptsize$\pm$2.5} \\
solo\_strong & None & 51.7\,{\scriptsize$\pm$2.9} & \cellcolor[RGB]{249,238,229}52.5\,{\scriptsize$\pm$5.0} & \cellcolor[RGB]{225,228,228}49.2\,{\scriptsize$\pm$1.4} & \cellcolor[RGB]{246,230,222}53.3\,{\scriptsize$\pm$5.2} & 73.1\,{\scriptsize$\pm$1.5} & \cellcolor[RGB]{241,214,207}76.4\,{\scriptsize$\pm$3.1} & \cellcolor[RGB]{247,233,225}74.4\,{\scriptsize$\pm$2.5} & \cellcolor[RGB]{240,210,203}76.8\,{\scriptsize$\pm$1.7} & 28.5\,{\scriptsize$\pm$0.9} & \cellcolor[RGB]{243,220,212}31.2\,{\scriptsize$\pm$4.9} & \cellcolor[RGB]{233,188,184}34.4\,{\scriptsize$\pm$0.9} & \cellcolor[RGB]{249,241,231}29.0\,{\scriptsize$\pm$8.4} \\
\midrule[\heavyrulewidth]
\multirow{4}{*}{onlyweak} & None & \multirow{4}{*}{21.7\,{\scriptsize$\pm$5.2}} & \cellcolor[RGB]{251,246,236}21.7\,{\scriptsize$\pm$9.5} & \cellcolor[RGB]{233,234,230}20.0\,{\scriptsize$\pm$4.3} & \cellcolor[RGB]{215,222,225}18.3\,{\scriptsize$\pm$6.3} & \multirow{4}{*}{52.9\,{\scriptsize$\pm$3.5}} & \cellcolor[RGB]{151,178,204}43.4\,{\scriptsize$\pm$5.3} & \cellcolor[RGB]{205,214,221}48.5\,{\scriptsize$\pm$5.6} & \cellcolor[RGB]{151,178,204}43.4\,{\scriptsize$\pm$4.0} & \multirow{4}{*}{1.1\,{\scriptsize$\pm$0.9}} & \cellcolor[RGB]{245,242,234}0.5\,{\scriptsize$\pm$0.9} & \cellcolor[RGB]{245,242,234}0.5\,{\scriptsize$\pm$0.9} & \cellcolor[RGB]{248,236,227}2.2\,{\scriptsize$\pm$0.9} \\
 & Collab. &  & \cellcolor[RGB]{241,214,207}25.0\,{\scriptsize$\pm$4.3} & \cellcolor[RGB]{225,228,228}19.2\,{\scriptsize$\pm$6.3} & \cellcolor[RGB]{243,221,214}\textbf{24.2}\,{\scriptsize$\pm$1.4} &  & \cellcolor[RGB]{231,183,179}\textbf{59.3}\,{\scriptsize$\pm$4.1} & \cellcolor[RGB]{245,226,218}\textbf{54.9}\,{\scriptsize$\pm$6.1} & \cellcolor[RGB]{187,202,216}46.8\,{\scriptsize$\pm$5.1} &  & \cellcolor[RGB]{239,238,232}0.0\,{\scriptsize$\pm$0.0} & \cellcolor[RGB]{245,242,234}0.5\,{\scriptsize$\pm$0.9} & \cellcolor[RGB]{251,246,236}1.1\,{\scriptsize$\pm$0.9} \\
 & Conform. &  & \cellcolor[RGB]{241,214,207}25.0\,{\scriptsize$\pm$2.5} & \cellcolor[RGB]{242,240,233}20.8\,{\scriptsize$\pm$1.4} & \cellcolor[RGB]{207,216,222}17.5\,{\scriptsize$\pm$2.5} &  & \cellcolor[RGB]{247,233,225}54.2\,{\scriptsize$\pm$2.3} & \cellcolor[RGB]{176,195,212}45.8\,{\scriptsize$\pm$6.7} & \cellcolor[RGB]{125,160,196}39.1\,{\scriptsize$\pm$5.9} &  & \cellcolor[RGB]{251,246,236}\textbf{1.1}\,{\scriptsize$\pm$0.9} & \cellcolor[RGB]{239,238,232}0.0\,{\scriptsize$\pm$0.0} & \cellcolor[RGB]{245,242,234}0.5\,{\scriptsize$\pm$0.9} \\
 & Compet. &  & \cellcolor[RGB]{233,189,185}\textbf{27.5}\,{\scriptsize$\pm$2.5} & \cellcolor[RGB]{231,181,178}\textbf{28.3}\,{\scriptsize$\pm$1.4} & \cellcolor[RGB]{163,186,208}13.3\,{\scriptsize$\pm$3.8} &  & \cellcolor[RGB]{125,160,196}22.6\,{\scriptsize$\pm$10.5} & \cellcolor[RGB]{125,160,196}28.3\,{\scriptsize$\pm$1.7} & \cellcolor[RGB]{218,224,226}\textbf{49.8}\,{\scriptsize$\pm$3.1} &  & \cellcolor[RGB]{245,242,234}0.5\,{\scriptsize$\pm$0.9} & \cellcolor[RGB]{248,236,227}\textbf{2.2}\,{\scriptsize$\pm$2.5} & \cellcolor[RGB]{234,193,189}\textbf{6.5}\,{\scriptsize$\pm$3.2} \\
\midrule
\multirow{4}{*}{weakstrong} & None & \multirow{4}{*}{48.3\,{\scriptsize$\pm$1.4}} & \cellcolor[RGB]{241,213,206}51.7\,{\scriptsize$\pm$2.9} & \cellcolor[RGB]{241,213,206}51.7\,{\scriptsize$\pm$2.9} & \cellcolor[RGB]{246,229,221}50.0\,{\scriptsize$\pm$4.3} & \multirow{4}{*}{80.5\,{\scriptsize$\pm$1.5}} & \cellcolor[RGB]{249,240,231}81.1\,{\scriptsize$\pm$5.9} & \cellcolor[RGB]{169,190,210}72.7\,{\scriptsize$\pm$3.0} & \cellcolor[RGB]{251,246,236}80.5\,{\scriptsize$\pm$1.5} & \multirow{4}{*}{17.2\,{\scriptsize$\pm$4.9}} & \cellcolor[RGB]{246,230,222}18.8\,{\scriptsize$\pm$4.7} & \cellcolor[RGB]{238,204,198}\textbf{21.5}\,{\scriptsize$\pm$2.4} & \cellcolor[RGB]{223,227,227}14.5\,{\scriptsize$\pm$4.3} \\
 & Collab. &  & \cellcolor[RGB]{198,210,219}43.3\,{\scriptsize$\pm$5.2} & \cellcolor[RGB]{214,128,130}62.5\,{\scriptsize$\pm$4.3} & \cellcolor[RGB]{243,221,214}50.8\,{\scriptsize$\pm$1.4} &  & \cellcolor[RGB]{238,204,198}84.8\,{\scriptsize$\pm$2.0} & \cellcolor[RGB]{234,193,188}\textbf{85.9}\,{\scriptsize$\pm$5.6} & \cellcolor[RGB]{246,229,221}82.2\,{\scriptsize$\pm$1.5} &  & \cellcolor[RGB]{240,238,232}16.1\,{\scriptsize$\pm$3.2} & \cellcolor[RGB]{245,242,234}16.7\,{\scriptsize$\pm$1.8} & \cellcolor[RGB]{228,231,229}\textbf{15.1}\,{\scriptsize$\pm$3.4} \\
 & Conform. &  & \cellcolor[RGB]{243,240,233}47.5\,{\scriptsize$\pm$0.0} & \cellcolor[RGB]{243,221,214}50.8\,{\scriptsize$\pm$3.8} & \cellcolor[RGB]{220,148,148}\textbf{58.3}\,{\scriptsize$\pm$5.8} &  & \cellcolor[RGB]{243,219,212}83.2\,{\scriptsize$\pm$3.5} & \cellcolor[RGB]{236,236,231}79.1\,{\scriptsize$\pm$1.2} & \cellcolor[RGB]{218,224,226}77.4\,{\scriptsize$\pm$4.1} &  & \cellcolor[RGB]{244,225,217}\textbf{19.4}\,{\scriptsize$\pm$4.3} & \cellcolor[RGB]{161,184,207}8.6\,{\scriptsize$\pm$0.9} & \cellcolor[RGB]{212,219,223}13.4\,{\scriptsize$\pm$3.7} \\
 & Compet. &  & \cellcolor[RGB]{214,128,130}\textbf{63.3}\,{\scriptsize$\pm$7.6} & \cellcolor[RGB]{214,128,130}\textbf{74.2}\,{\scriptsize$\pm$3.8} & \cellcolor[RGB]{243,221,214}50.8\,{\scriptsize$\pm$5.8} &  & \cellcolor[RGB]{233,187,183}\textbf{86.5}\,{\scriptsize$\pm$0.6} & \cellcolor[RGB]{245,226,218}82.5\,{\scriptsize$\pm$4.6} & \cellcolor[RGB]{238,204,198}\textbf{84.8}\,{\scriptsize$\pm$1.0} &  & \cellcolor[RGB]{223,227,227}14.5\,{\scriptsize$\pm$2.8} & \cellcolor[RGB]{228,231,229}15.1\,{\scriptsize$\pm$3.3} & \cellcolor[RGB]{178,196,213}10.2\,{\scriptsize$\pm$1.9} \\
\midrule
\multirow{4}{*}{onlystrong} & None & \multirow{4}{*}{45.0\,{\scriptsize$\pm$5.0}} & \cellcolor[RGB]{214,128,130}\textbf{61.7}\,{\scriptsize$\pm$2.9} & \cellcolor[RGB]{214,128,130}\textbf{60.8}\,{\scriptsize$\pm$5.8} & \cellcolor[RGB]{214,128,130}59.2\,{\scriptsize$\pm$8.0} & \multirow{4}{*}{63.0\,{\scriptsize$\pm$2.3}} & \cellcolor[RGB]{205,214,221}58.6\,{\scriptsize$\pm$6.3} & \cellcolor[RGB]{214,128,130}76.8\,{\scriptsize$\pm$2.7} & \cellcolor[RGB]{247,233,225}64.3\,{\scriptsize$\pm$4.2} & \multirow{4}{*}{33.9\,{\scriptsize$\pm$2.8}} & \cellcolor[RGB]{249,241,232}34.4\,{\scriptsize$\pm$2.5} & \cellcolor[RGB]{246,230,222}35.5\,{\scriptsize$\pm$4.3} & \cellcolor[RGB]{206,215,222}29.6\,{\scriptsize$\pm$2.5} \\
 & Collab. &  & \cellcolor[RGB]{214,128,130}59.2\,{\scriptsize$\pm$1.4} & \cellcolor[RGB]{230,180,177}51.7\,{\scriptsize$\pm$8.0} & \cellcolor[RGB]{214,128,130}\textbf{60.8}\,{\scriptsize$\pm$1.4} &  & \cellcolor[RGB]{222,154,153}\textbf{72.4}\,{\scriptsize$\pm$1.2} & \cellcolor[RGB]{214,128,130}82.8\,{\scriptsize$\pm$1.7} & \cellcolor[RGB]{212,219,224}59.3\,{\scriptsize$\pm$3.8} &  & \cellcolor[RGB]{248,236,227}34.9\,{\scriptsize$\pm$3.7} & \cellcolor[RGB]{244,225,217}36.0\,{\scriptsize$\pm$3.3} & \cellcolor[RGB]{217,223,225}30.6\,{\scriptsize$\pm$1.6} \\
 & Conform. &  & \cellcolor[RGB]{220,148,148}55.0\,{\scriptsize$\pm$7.5} & \cellcolor[RGB]{214,128,130}57.5\,{\scriptsize$\pm$6.6} & \cellcolor[RGB]{225,164,163}53.3\,{\scriptsize$\pm$3.8} &  & \cellcolor[RGB]{248,236,227}64.0\,{\scriptsize$\pm$2.1} & \cellcolor[RGB]{214,128,130}\textbf{87.2}\,{\scriptsize$\pm$2.1} & \cellcolor[RGB]{180,197,213}56.2\,{\scriptsize$\pm$3.3} &  & \cellcolor[RGB]{246,230,222}35.5\,{\scriptsize$\pm$4.3} & \cellcolor[RGB]{243,220,213}\textbf{36.6}\,{\scriptsize$\pm$6.5} & \cellcolor[RGB]{206,215,222}29.6\,{\scriptsize$\pm$3.4} \\
 & Compet. &  & \cellcolor[RGB]{233,234,230}43.3\,{\scriptsize$\pm$7.6} & \cellcolor[RGB]{230,180,177}51.7\,{\scriptsize$\pm$5.2} & \cellcolor[RGB]{223,156,155}54.2\,{\scriptsize$\pm$1.4} &  & \cellcolor[RGB]{229,177,174}70.0\,{\scriptsize$\pm$2.5} & \cellcolor[RGB]{214,128,130}83.8\,{\scriptsize$\pm$1.0} & \cellcolor[RGB]{214,128,130}\textbf{83.5}\,{\scriptsize$\pm$1.2} &  & \cellcolor[RGB]{244,225,217}\textbf{36.0}\,{\scriptsize$\pm$3.4} & \cellcolor[RGB]{251,246,236}33.9\,{\scriptsize$\pm$1.6} & \cellcolor[RGB]{246,230,222}\textbf{35.5}\,{\scriptsize$\pm$3.2} \\
\bottomrule
\end{tabular}%
}
\vspace{-1mm}
\caption{Main results (avg@3 success rate, \%) on the test set. Cell color shows the change from \emph{Pre}: \colorbox[RGB]{229,175,172}{rose} for gains and \colorbox[RGB]{175,194,212}{blue} for drops (\emph{Pre} : same team before evolution), darker for larger changes. \textbf{Bold}: best structure of interdependence for each team, mechanism, and benchmark. Values are mean\,$\pm$\,SD over three rollouts.}
\vspace{-5mm}
\label{tab:main}
\end{table*}

\paragraph{Shared settings.}
In every condition, the end-of-iteration decision is made by the chairman's own model with a fixed decision prompt (see Appendix~\ref{app:prompts} for the prompt), so no model outside the team decides for it.\footnote{The only components outside the team are the L2 contradiction judge and the SkillOpt and MemRL update models, all fixed to the strong model.} Gate sessions run under the same condition, tiers, and chairman as training, and all conditions, including none, share the same L3 gate (Figure~\ref{fig:AGORA}, \textit{`L3 Gate'}) and revision cap, so that only the social layer differs. Appendix~\ref{app:interdependence} gives the full specification of each condition.

\section{Experimental Setup}
\label{experiment}

We cross every evolution mechanism with every combination of team composition and interdependence structure on three benchmarks, comparing each evolved team with its unevolved self and a single agent of each tier (RQ1, Section~\ref{sec:results}), and then trace its final-evaluation sessions to see how it reorganizes and why (RQ2 and RQ3, Section~\ref{sec:analysis}).

\paragraph{Benchmarks}
We use web-based reasoning (GAIA; \citealp{mialon2024gaia}), long codebase understanding (LoCoBench; \citealp{qiu2025locobench}), and document-based numerical question answering (OfficeQA Pro; \citealp{opsahlong2026officeqa}), each split into non-overlapping train, val, and test sets (split sizes and task formats in Appendix~\ref{app:benchmarks}).

\paragraph{Models}
The weak and strong tiers are \textit{Qwen3-30B-A3B-Instruct-2507} and \textit{Qwen3.6-35B-A3B}, run with thinking off and temperature 0.1 (model selection in Appendix~\ref{app:team_config}; serving in Appendix~\ref{app:infra}). 

\paragraph{Session Budgets}
Each session is limited to 900 seconds and 100 team messages (50 for solo), and each agent to 50 steps. A session terminated by its time or message limit is not graded and counts as a failure (0) in every score, including both gates; only sessions that crash are excluded. 

\paragraph{Experiment Matrix}
\label{par:experiment_matrix}
We use two kinds of baselines. Solo baselines give the whole task to one weak or strong agent with every tool. Preevolve baselines evaluate each of the five configurations (two solo setups and three team compositions) on the test set without evolution. Each benchmark thus comprises 47 experiments: five preevolve baselines, six solo runs, and 36 MAS runs. 

\paragraph{Evaluation}
Tasks are scored as success or failure: exact match for GAIA, an LCBS score of at least 3.0 for LoCoBench, and a verdict from \textit{gpt-5-mini} for OfficeQA Pro, a judge from a different model family to avoid self-preference.
\section{Does Social Scaffolding Change What a Team Achieves?}
\label{sec:results}
To answer RQ1, we compare each team before evolution, after evolving under \emph{none}, and after evolving under structure of interdependence (Table~\ref{tab:main}).


\paragraph{Evolution alone falls short.}
Evolution alone brings little across the 27 MAS settings: it raises the score by at most 4.3 points over \emph{Pre} in 23 of them and even lowers it in 11, and 17 of them fall short of a single strong agent evolved with the same mechanism. Large gains appear mainly in teams of identical strong agents (13.8--16.7 points), the most expensive teams to deploy.

\paragraph{The right scaffolding makes evolution pay off.}
An appropriate structure outperforms evolution alone in 24 of the 27 settings, and the best combination of team composition and structure surpasses the strong agent evolved alone on every benchmark and mechanism, by up to 25.0 points (74.2 vs.\ 49.2 with SkillOpt on GAIA). The same evolution mechanism thus yields far larger gains once the team is organized appropriately.

\paragraph{The best scaffolding differs from team to team.}
Which structure helps depends on who is in the team. On GAIA with SkillOpt, the onlystrong team is best under evolution alone (60.8), but the weakstrong team is best under competition (74.2). On LoCoBench with Meta-Team, competition lifts the onlystrong team (70.0) but collapses the onlyweak team (22.6). Gains are large both when members clearly differ in ability and when all are strong, and no structure helps when one member is near zero: on OfficeQA Pro, where the weak model scores 2.7 alone, even the best mixed team (21.5) stays below the strong agent alone (28.5). Section~\ref{sec:analysis} explains why: what the right scaffolding changes is who leads the team.

\begin{table*}[t]
\centering
\setlength{\tabcolsep}{3.5pt}
\resizebox{\textwidth}{!}{%
\begin{tabular}{ll rrrr @{\hspace{1.6em}} l rrrr @{\hspace{1.6em}} l rrrr}
\toprule
\multicolumn{6}{c}{\textbf{GAIA weakstrong}} &
\multicolumn{5}{c}{\textbf{LoCoBench weakstrong}} &
\multicolumn{5}{c}{\textbf{LoCoBench onlystrong}} \\
\multicolumn{6}{c}{} &
\multicolumn{5}{c}{\small chair: planner (W)} &
\multicolumn{5}{c}{\small chair: planner (S)} \\
\cmidrule(r{1.2em}){1-6} \cmidrule(r{1.2em}){7-11} \cmidrule{12-16}
Run & Chair & Agents & Calls & TO & avg@3 &
Run & Agents & Calls & TO & avg@3 &
Run & Agents & Calls & TO & avg@3 \\
\midrule
Solo strong, \emph{Pre} & -- & 1.00 & 22.6 & 0.0 & 51.7 &
Solo strong, \emph{Pre} & 1.00 & 24.3 & 12.8 & 73.1 &
Team, \emph{Pre} & 3.20 & 73.3 & 23.2 & 63.0 \\
Team, \emph{Pre} & plan\_agent (W) & 2.76 & 46.8 & 7.5 & 48.3 &
Team, \emph{Pre} & 4.18 & 80.0 & 5.4 & 80.5 &
SO, none & 1.40 & 27.0 & 12.8 & 76.8 \\
SO, none & plan\_agent (W) & 2.97 & 39.6 & 3.3 & 51.7 &
SO, conform. & 4.62 & 107.3 & 5.1 & 79.1 &
SO, collab. & 1.00 & 19.7 & 5.1 & 82.8 \\
SO, compet. & web\_agent (S) & 1.38 & 19.4 & 0.0 & \textbf{74.2} &
SO, compet. & 4.55 & 91.5 & 3.4 & 82.5 &
SO, conform. & 1.00 & 18.3 & 4.4 & \textbf{87.2} \\
MT, compet. & web\_agent (S) & 1.01 & 17.0 & 0.8 & 63.3 &
MR, conform. & 4.43 & 100.9 & 6.1 & 77.4 &
SO, compet. & 1.01 & 21.1 & 3.7 & 83.8 \\
MR, compet. & plan\_agent (W) & 2.98 & 40.0 & 7.5 & 50.8 &
MR, compet. & 4.01 & 74.8 & 4.7 & \textbf{84.8} &
MR, conform. & 1.17 & 29.3 & 11.8 & 56.2 \\
\bottomrule
\end{tabular}%
}
\caption{Process metrics per session in the final evaluation for the runs discussed in Section~\ref{sec:analysis} (GAIA $n{=}120$, LoCoBench $n{=}297$ sessions per run; all scores in Table~\ref{tab:main}).  \emph{Chair}: the role holding the chair and its model; on LoCoBench, the chair stays the planner in every run shown. \emph{Agents}: roles started in the session; \emph{Calls}: LLM calls; \emph{TO}: share of sessions stopped by the 900-second limit (\%). Bold: best avg@3 in each block. (SO: SkillOpt, MT: Meta-Team, MR: MemRL; W: weak, S: strong)}
\vspace{-4mm}
\label{tab:process}
\end{table*}

\section{How the Team Reorganizes, and Why It Works}
\label{sec:analysis}
To answer RQ2 and RQ3, we trace the final-evaluation sessions of the teams with the largest effects (Table~\ref{tab:process}) and relate them to human groups. Sections~\ref{sec:who} and~\ref{sec:coordination} answer RQ2, and Section~\ref{sec:grown} answers RQ3. Additional experiments with a much weaker model, different assignments of strong roles, and other model families lead to largely the same conclusions (Appendix~\ref{app:analysis_results}).

\subsection{Authority Moves to the Most Capable Member}
\label{sec:who}
The first way the team reorganizes (RQ2) is in who leads it. GAIA weakstrong starts with the weak plan\_agent as chair, which relays sub-tasks to the strong web\_agent and doubles the LLM calls of the strong agent alone (46.8 vs.\ 22.6). Competition removes this relay by handing the chair to the strong web\_agent under SkillOpt and Meta-Team, which then works alone at solo-level cost (74.2 and 63.3). Promoting a weak role instead backfires: on OfficeQA Pro weakstrong, competition hands the chair to a weak role and drops the team below evolution alone (15.1 vs.\ 21.5 with SkillOpt; Table~\ref{tab:main}). Every chair change under competition, with the score that decided it, is listed in Table~\ref{tab:promotions}. Like human groups, which gain from their most capable member only when they recognize it~\cite{littlepage1997effects} and pay for poorly suited leaders~\cite{milgram1963behavioral,janis1972victims}, competition helps only when the right role rises.

\subsection{The Others Step Back, Unless They Remove a Bottleneck}
\label{sec:coordination}
The second way the team reorganizes (RQ2) is in who works. Coordination pays off only when it removes a bottleneck. On LoCoBench weakstrong, the weak readers take over the reading, halving the timeouts of the strong agent alone (5.4\% vs.\ 12.8\%) and outscoring it before evolution (80.5 vs.\ 73.1; Table~\ref{tab:process}). Coordination beyond this costs the team: conformity raises the LLM calls per session from 80.0 to 107.3 with SkillOpt and 100.9 with MemRL, and drops the team below \emph{Pre} (79.1 and 77.4), adding cost without removing any further bottleneck.

In onlystrong, where every member is equally capable, splitting the work adds waiting (23.2\% timeouts), so SkillOpt moves the work to the planner alone (1.00--1.40 active agents). Here conformity helps most, with the fewest LLM calls (18.3): it writes a planner-implements rule and removes the developer within two iterations, as conformity spreads one practice and preserves it~\cite{henrich1998evolution}. The same structure thus helps when it removes coordination and hurts when it adds it. Unlike humans, who may loaf beside more capable members~\cite{latane1979many,karau1993social}, the others drop out, and the team does better.

\subsection{The Team Grows Its Leader}
\label{sec:grown}
\begin{table}[!htbp]
\centering
\small
\begin{tabular*}{\columnwidth}{@{\extracolsep{\fill}}lcccc@{}}
\toprule
 & \multicolumn{2}{c}{GAIA} & \multicolumn{2}{c}{LoCoBench} \\
\cmidrule(lr){2-3} \cmidrule(lr){4-5}
 & Solo & Team & Solo & Team \\
\midrule
avg@3 & 49.2 & \textbf{74.2} & 74.4 & \textbf{87.2} \\
Skills accepted & 0/4 & 2/4 & 2/4 & 2/4 \\
Solved only here & 5 & \textbf{35} & 8 & \textbf{46} \\
\bottomrule
\end{tabular*}
\caption{Team-grown agent (\emph{Team}) vs.\ the same model alone (\emph{Solo}) under SkillOpt in GAIA weakstrong (competition) and LoCoBench onlystrong (conformity). Skills accepted: this agent's skills that passed the L1 gate, out of its candidates that reached the gate. Solved only here: (task, rollout) pairs only this agent solved. 
}
\vspace{-3mm}
\label{tab:grown_solo}
\end{table}

The reorganized team outperforms the same model working alone (RQ3) because the agent left leading it has learned from its teammates. This agent outperforms the same model evolved alone (Table~\ref{tab:grown_solo}), solving 35 (task, rollout) pairs the solo agent misses against 5 the other way on GAIA, and 46 against 8 on LoCoBench. On GAIA, the strong agent alone accepts none of its four candidate skills, whereas web\_agent accepts two of its four within the team, and on LoCoBench the planner follows the workflow its team wrote into the constitution. MemRL under conformity consolidates the work without such learning and falls below \emph{Pre} (56.2 vs.\ 63.0). The team thus acts as scaffolding in its original sense~\cite{wood1976role,vygotsky1978mind}: support the leader needs while it develops and no longer needs once it has.

\subsection{What This Means for Team Design}
\label{sec:design}
Taken together, the answers to RQ1--RQ3 suggest three principles for designing a self-evolving MAS. First, decide who leads before deciding how agents learn: on GAIA, the same weakstrong team scores 74.2 with the strong web\_agent in the chair but at most 62.5 while the weak plan\_agent keeps it. Second, let performance choose the leader, but check that it picks a capable role: competition helps when it promotes the strong web\_agent and hurts when it promotes a weak role, as on OfficeQA Pro (Section~\ref{sec:who}). Third, keep teammates only where they remove a bottleneck: readers that share a long codebase cut timeouts on LoCoBench, whereas in an onlystrong team the work moves to the planner alone and the score rises (Section~\ref{sec:coordination}).
\section{Conclusion}
\label{sec:conclusion}
We investigate whether the social layer of a self-evolving MAS shapes its evolution, as in human groups, and build AGORA to vary team composition and the structure of interdependence with all else fixed. Evolution alone rarely helps; the right scaffolding works by reorganizing the team: authority moves to the most capable member, the others step back unless they remove a bottleneck, and the member left leading, having learned from its teammates, outperforms the same model alone. An evolution mechanism alone is not enough, since the social scaffolding under which a team evolves decides how its agents learn and who gets to lead.
\section*{Limitations}
Results are partly confounded with benchmarks, so differences in task
type may contribute to the patterns we attribute to ability. Our experiments that change the
models (Appendix~\ref{app:analysis_results}) partially separate the two, but a
full factorial design over ability and task type is left to future work.
Each session has a 900-second wall-clock limit, and up to six sessions run concurrently on shared GPUs. Because more active agents can raise the server load and slow down each call, the observation that the team improves when the others step back may partly reflect this serving setup.
Due to cost, we run evolution only once per condition. The reported standard deviations reflect only the noise of evaluating the same evolved team three times, not the variance of the evolution process itself.
\section*{Acknowledgments}
This work was supported by the Institute of Information \& Communications Technology Planning \& Evaluation (IITP) grant funded by the Korea government (MSIT) (No.~2021-0-01341, Artificial Intelligence Graduate School Program (Chung-Ang University)), the National Research Foundation of Korea (NRF) grant funded by the Korea government (MSIT) (No.~RS-2026-25494299), and the ``Advanced GPU Utilization Support Program'' funded by the Government of the Republic of Korea (Ministry of Science and ICT).
\bibliography{custom}
\clearpage
\appendix

\section{Benchmark Details}\label{app:benchmarks}
For each benchmark, we fix 20 val problems and divide the remaining problems into train and test at a 4:5 ratio (at least 40 test problems, seed 42). Gates always use val, and final evaluation always uses test. Training tasks are drawn epoch-wise: the train problems are shuffled once per epoch and consumed in consecutive, non-overlapping batches. The last batch of an epoch may be smaller, after which the problems are reshuffled. Split sizes and scoring criteria are summarized in Table~\ref{tab:benchmarks}.

For GAIA, we use only the 77 problems without attached files, and answers are scored by exact match on the text after ``FINAL ANSWER:''. Web search uses the Serper API, with identical queries cached for reproducibility, and each task starts with an empty working directory. Since train has only 17 problems, the same 17 problems are repeated in a reshuffled order with the same seeds over the 4 iterations, so per-iteration batch success rates on GAIA are compared on the same problems.

For LoCoBench, we use 199 Python scenarios. Because the whole codebase has to be read, a 200k-token context is required. A task counts as a success if its LCBS score, which is based on static analysis and ranges up to 5, is 3.0 or higher. Over the 4 iterations, each of the 80 train problems is solved once, so per-iteration batch success rates also reflect differences in problem difficulty.

For OfficeQA, the Pro set consists of 133 hard problems. One to three relevant documents are copied into each task's working directory, and an LLM judge (openai/gpt-5-mini) compares the answer with the ground truth and returns correct or incorrect.

\begin{table}[ht]
\centering
\large
\resizebox{\columnwidth}{!}{%
\begin{tabular}{lcccll}
\toprule
Benchmark & Train & Val & Test & Success criterion & Judge \\
\midrule
GAIA & 17 & 20 & 40 & Exact match of the final answer & -- \\
LoCoBench & 80 & 20 & 99 & LCBS $\geq$ 3.0 (out of 5) & -- \\
OfficeQA Pro & 51 & 20 & 62 & Judged correct & gpt-5-mini \\
\bottomrule
\end{tabular}%
}
\caption{Benchmark splits and scoring. The validation set (20 tasks) is fixed first; the rest is split train : test = 4 : 5 (test $\geq$ 40, seed 42). Gates always use the validation set and the final evaluation always uses the test set.}
\label{tab:benchmarks}
\end{table}

\section{Team Configurations}\label{app:team_config}
Table~\ref{tab:roles} lists the roles of each benchmark, the tools assigned to each role, and which role is strong in weakstrong. In solo, a single agent uses all tools of the benchmark and has no constitution. Since it has no teammates, it evolves through L1 alone, without L2 or L3.

\begin{table*}[t]
\centering
\small
\resizebox{\textwidth}{!}{%
\begin{tabular}{llll}
\toprule
Benchmark & Role & Tools & Model in \textit{weakstrong} \\
\midrule
GAIA & plan\_agent (chairman) & -- & weak \\
     & web\_agent & web\_search, web\_fetch, bash, read\_file & \textbf{strong} \\
     & file\_agent & bash, read\_file, write\_file & weak \\
     & answer\_agent & bash, read\_file & weak \\
\midrule
LoCoBench & planner (chairman) & read\_file, bash & weak \\
          & developer & read\_file, write\_file, bash & \textbf{strong} \\
          & reader-1--4 & read\_file, bash & weak \\
\midrule
OfficeQA & planner (chairman) & -- & weak \\
         & reader-1, reader-2 & bash, read\_file, web\_search, web\_fetch & \textbf{strong} \\
         & calculator & bash & weak \\
         & answer\_agent & -- & weak \\
\bottomrule
\end{tabular}%
}
\caption{Roles, tools, and model assignment in \textit{weakstrong} for each benchmark. Weak: Qwen3-30B-A3B-Instruct-2507; strong: Qwen3.6-35B-A3B. All roles also have the messaging tools; the chairman additionally has the role-management (start\_agent / stop\_agent) and answer-submission (finalize\_task) tools. In \textit{weakstrong}, only the bottleneck role is strong; OfficeQA runs this role (reader) as two identical instances, both strong.}
\label{tab:roles}
\end{table*}

We chose the weak and strong models based on the MMLU-Pro, GPQA, and SWE-bench Verified scores reported on their official model cards, although the evaluation settings there may differ from ours. The weak model, Qwen3-30B-A3B-Instruct-2507, reports 78.4 on MMLU-Pro and 70.4 on GPQA; it does not report SWE-bench Verified (43.2 on LiveCodeBench v6). The strong model, Qwen3.6-35B-A3B, reports 85.2 on MMLU-Pro, 86.0 on GPQA (Diamond), and 73.4 on SWE-bench Verified. Both are MoE models with about 3B active parameters, and thinking mode is turned off for both. In weakstrong, the strong role is the bottleneck role of each benchmark, that is, the role that gathers information or writes the final output. When that role has multiple instances, all of them are made strong.

All template roles use a temperature of 0.1 and a limit of 50 steps per agent, and each task session has a time limit of 900 seconds, separate from the time budget for reflection (Appendix~\ref{app:engine}). The maximum number of output tokens is 16,384 for GAIA (4,096 for answer\_agent), 10,000 for LoCoBench, and 16,384 for OfficeQA (8,192 for the readers and the calculator, and 4,096 for answer\_agent). Solo agents use 16,384 on every benchmark. The team message limit is 100 for MAS and 50 for solo, and a message sent to several recipients counts once per recipient. Representative role prompts and the full constitutions are shown in Appendix~\ref{app:prompts}.

\section{Differences from Meta-Team}
\label{app:metateam}

AGORA builds on the public implementation of Meta-Team~\citep{hao2026evolve}. From it, AGORA inherits the basic loop: a chairman-led team of LLM agents solves tasks by exchanging messages, evolves only its external state rather than model weights, and revises itself at the individual (L1), relational (L2), and team (L3) levels. Table~\ref{tab:metateam} summarizes what AGORA changes.

The main change is in purpose. Meta-Team uses the three-level loop to improve a single team under one fixed social design: the chairman alone proposes and applies team changes, while other members can only send suggestions. At L3, this design is closest to our conformity condition. AGORA instead treats this social layer as the variable under study. It parameterizes L2 and L3 by four structures of interdependence, adds team composition as a second axis, and makes L1 a slot for interchangeable evolution mechanisms so that their effect can be separated from that of social scaffolding.

Two further changes keep conditions comparable. First, Meta-Team applies team changes immediately, with no check on held-out data; AGORA checks SkillOpt skill revisions and every L3 revision on the same validation set (Section~\ref{sec:gate}), so that no condition keeps changes that clearly hurt the team. Second, Meta-Team solves training tasks one at a time and reflects right after each task; AGORA runs sessions concurrently with explicit budgets and defers reflection until the batch is done (Appendix~\ref{app:engine}). Under the Meta-Team mechanism, we keep the original sequential execution and single training pass.

\begin{table}[h]
\centering
\resizebox{\columnwidth}{!}{%
\begin{tabular}{lll}
\toprule
 & \textbf{Meta-Team} & \textbf{AGORA} \\
\midrule
Goal & Improve one team & Compare social conditions \\
L1 update & Agent's own edits & Meta-Team, SkillOpt, or MemRL \\
L2 & Single fixed protocol & Set by structure; off under None \\
L3 proposal & Chairman alone; others suggest & Tier~1 proposes \\
L3 decision & Applied by the chairman & Rule set by structure \\
Chairman & Fixed by default & Fixed, or rotates under Competition \\
Team composition & One model for all roles & Weak/strong per role \\
Held-out gate & None; changes apply immediately & L1 (SkillOpt) and L3, same val set \\
Execution & One task at a time & Concurrent sessions$^\dagger$ \\
Reflection & Right after each task & Deferred until the batch ends$^\dagger$ \\
\bottomrule
\end{tabular}%
}
\caption{What AGORA inherits from and changes in Meta-Team. $^\dagger$Except under the Meta-Team mechanism, which keeps the original sequential execution and single training pass (Table~\ref{tab:mechanisms}).}
\label{tab:metateam}
\end{table}

\section{Evolution Details}\label{app:evolution}
This section follows the three evolution levels in Table~\ref{tab:layers}: the L1 mechanisms that update each role (Appendix~\ref{app:mechanisms}), the relational notes of L2, the team-level proposals of L3, and the gates that decide whether L1 and L3 changes are applied (Appendix~\ref{app:gate}).

\subsection{Models Used at Each Stage}\label{app:model_use}
Table~\ref{tab:model_use} lists the model that performs each stage. Everything an agent does inside a session, including its L1, L2, and L3 reflection, uses that agent's own model, so under Meta-Team the team's own models revise themselves. In contrast, the SkillOpt optimizer, the MemRL memory writer, and the L2 contradiction judge are fixed to the strong model regardless of team composition. In the logs of all GAIA Meta-Team runs, every L1 tool call (an edit or a skip) was made by the calling agent's own model.

\begin{table}[htbp]
\centering
\large
\resizebox{\columnwidth}{!}{%
\begin{tabular}{ll}
\toprule
Stage & Model \\
\midrule
Task execution & the role's own model (weak or strong) \\
L1, L2, and L3 reflection in a session & the agent's own model \\
L3 decision at the end of an iteration & the current chairman's model \\
Gate sessions & as in task execution \\
SkillOpt analysis and edits & the strong model (fixed) \\
MemRL memory writing & the strong model (fixed) \\
L2 contradiction judge & the strong model (fixed) \\
MemRL retrieval embedding & bge-m3 \\
OfficeQA scoring & openai/gpt-5-mini \\
\bottomrule
\end{tabular}%
}
\caption{Model used at each stage. ``Fixed'' means the same model regardless of team composition.}
\label{tab:model_use}
\end{table}

\subsection{L1: Evolution Mechanisms}\label{app:mechanisms}

\subsubsection{Meta-Team}
Meta-Team has no separate mechanism: right after a session, in the L1 phase, agents directly edit their own prompt patches and skills. There is no L1 gate, and as in the original design, batches run sequentially for a single iteration. L3 changes go through the same L3 gate as in the other mechanisms. The preevolve runs give Meta-Team zero iterations, so the template team is evaluated on test as is.

\subsubsection{SkillOpt}

We use the reflect $\rightarrow$ merge $\rightarrow$ apply functions of the original SkillOpt without modification and run them concurrently for each role. The input to the analysis model follows the original format: trajectories that pair each action with its outcome, the grading result appended at the end (prediction, ground truth, and correctness, or the LCBS sub-scores), the failure reason, and the system prompt of that session. Each analysis call batches up to eight trajectories from the same role. If a batch exceeds the context limit, it is split evenly into the fewest smaller batches that fit, and the longest tool outputs are truncated first only when a single session alone exceeds the limit.

Each iteration uses an edit budget of 4, 16 parallel workers, and the strong model as the optimizer model. Every reflect call writes to a fresh output directory so that revisions from previous iterations are not reused, and slightly malformed JSON in analysis responses is repaired with json\_repair, a dependency of the original implementation.

The resulting candidate skill documents are adopted only if they pass the L1 gate. The L1 gate uses the same 20-problem validation set and one-sided binomial test as the L3 gate (Section~\ref{sec:gate}), but whereas the L3 gate keeps a revision unless it is clearly worse, the L1 gate adopts a candidate only if it is clearly better than the current skill. For example, if the current skill solves 10 of the 20 problems, a candidate must solve at least 12 to be adopted. The significance level is $\alpha = 0.3$, the same as for the L3 gate. Because merely re-running the same team changes the outcome of about a third of the validation problems, a strict level would fail to distinguish genuine improvements from noise and would reject almost every candidate (full thresholds in Appendix~\ref{app:gate}).

\subsubsection{MemRL}
We use the query-based retrieval of the original MemRL runner. Only memories above the similarity threshold become candidates; they are ranked by $0.5\,z(\mathrm{sim}) + 0.5\,z(Q)$, and up to five are selected. With probability 10\%, the selection is random instead, with the random state fixed by the session seed, and memories with $Q < -0.8$ are excluded. All selected memories are added to the session prompt, each marked with its success or failure and its similarity. Every selected memory then receives a reward of $r = +1$ (success) or $-1$ (failure), and $Q$ is updated as $Q \leftarrow (1-\alpha)Q + \alpha r$ with $\alpha = 0.3$.

To create a new memory, the trajectory, including actions but excluding the system prompt that contains the retrieved older memories, is summarized as a procedural script. Failed trajectories are also stored, as ``scripts of failed attempts.'' The output is capped at 12,800 tokens and each memory at 12,000 characters. If the similarity to an existing retrieval key is 0.8 or higher, the memory is attached to that key; otherwise, it is indexed under a new key. MemRL has no gate.

We made five changes to the original. First, we use a local bge-m3 embedding model instead of the original text-embedding-3-large, and therefore re-tuned the threshold per benchmark so that about 16\% of val–train pairs pass, using only the unlabeled task inputs; no labels or scores were used: 0.38 for GAIA, 0.62 for LoCoBench, and 0.56 for OfficeQA (the original also uses a different value per benchmark, ranging from 0.25 to 0.62). Second, the retrieval key uses only the problem content, without the shared prompt template; with the template included, even unrelated GAIA problems reached a similarity of 0.64. Third, each role has its own memory bank, and every experiment starts with an empty bank. Fourth, updates are applied together at the end of each iteration rather than after each task, because sessions in a batch run concurrently. Finally, the model that writes memories is fixed to the strong model regardless of team composition.

\subsection{L2: Relational Notes}\label{app:l2}
L2 produces two artifacts: a profile of each teammate and collaboration notes for each teammate. A profile holds at most 5 strengths, 5 weaknesses, 3 notes, and 5 recorded contradictions, newest first, together with a cumulative observation count. Tier 1 can read the notes of the whole team, while Tier 2 can read only its own. Under collaboration and competition, a staged note is finalized once every teammate the author interacted with in the task has responded; if such a teammate finishes its L2 phase without responding, the note is finalized automatically.

To check for contradictions, a strong judge is asked whether a new observation contradicts earlier ones and answers either \textit{NO} or \textit{YES: <one sentence>}. A contradiction is only flagged and does not block saving. Since sessions in a batch run concurrently, if the file at save time differs from the version a session saw at its start, the notes written by other sessions are kept below the new note. In SkillOpt and MemRL, L2 notes are also passed to the L1 update as reference text.

\subsection{L3: Proposals and Review}\label{app:l3}
A new role is proposed as a config and prompt in a new folder, and a removal is proposed by marking the config as deleted. Proposals are not applied immediately but accumulate in a iteration buffer. Tier 2 agents cannot make proposals directly, but they can send a suggestion, which is added to the chairman's review list.

Under collaboration and competition, proposals go through peer review. A proposal passes once a majority ($\lfloor m/2 \rfloor + 1$) of the teammates who took part in the task, excluding the proposer, have commented on it. Without a majority, it still passes 180 seconds after its last edit, and teammates who did not comment are recorded as non-responses. Reviews are stored per version of each proposal file, so editing one proposal does not erase the reviews of another. However a session ends, whether by a timeout or by the chairman skipping, proposals not yet applied are passed on to the gate at the end of the iteration. As safeguards, the model of a new role is reset to the strong model in onlystrong teams and to the weak model otherwise, including weakstrong.

\subsection{Gates}\label{app:gate}
Both gates score candidates on the same 20 val problems, but they point in opposite directions: L1 accepts a change only if it is clearly better, whereas L3 accepts a change unless it is clearly worse. L1 follows the original SkillOpt, which keeps a new skill only if it scores higher than the current one, and adds a binomial test against the noise of the small val set. L3 instead checks structural changes that the team itself decided under its structure of interdependence, so it blocks only changes that clearly hurt and otherwise leaves the decision to the team. Both use $\alpha = 0.3$. Given the noise in 20 val problems (re-running the same team changes the outcome of about a third of the tasks), we use this relaxed threshold instead of a strict significance level. Table~\ref{tab:gate} lists the resulting thresholds for $n = 20$: L1 needs about two or more successes above the reference, and L3 reverts only at two or more successes below it. Gate sessions run under the same structure of interdependence, tiers, and chairman as training, except the shared iteration-0 L1 baseline, which is measured once under none (Appendix~\ref{app:l1_gate}).

\begin{table}[t]
\centering
\large
\resizebox{\columnwidth}{!}{%
\begin{tabular}{cccc}
\toprule
\multicolumn{2}{c}{Reference} & L1: accept if candidate & L3: revert if changed team \\
\cmidrule(lr){1-2}
Successes (/20) & Rate $b$ & $\geq$ (/20) & $\leq$ (/20) \\
\midrule
0 & 0.00 & 2 & -- \\
1 & 0.05 & 2 & -- \\
2 & 0.10 & 4 & 0 \\
3 & 0.15 & 5 & 1 \\
4 & 0.20 & 6 & 2 \\
5 & 0.25 & 7 & 3 \\
6 & 0.30 & 8 & 4 \\
7 & 0.35 & 9 & 5 \\
8 & 0.40 & 10 & 6 \\
9 & 0.45 & 11 & 7 \\
10 & 0.50 & 12 & 8 \\
11 & 0.55 & 13 & 9 \\
12 & 0.60 & 14 & 10 \\
\bottomrule
\end{tabular}%
}
\caption{Gate thresholds for $n=20$ and $\alpha=0.3$. The reference is the stored skill score for L1 and the current team's score for L3. L1 accepts a candidate skill only if it is clearly better (about two or more tasks above the reference); L3 reverts the changes only if the changed team is clearly worse (two or more tasks below). ``--'': never reverted.}
\label{tab:gate}
\end{table}

\subsubsection{L1 Gate (SkillOpt)}\label{app:l1_gate}
Let $b$ be the baseline success rate and $s$ the number of successes of the candidate out of $n$ tasks. With $p_0 = \max(b, 0.5/n)$ and $X \sim \mathrm{Bin}(n, p_0)$, the candidate is accepted $\iff s > b\,n \;\wedge\; P(X \ge s) \le \alpha$. Tasks in which the role was not called are excluded, so $n$ can be smaller than 20.

The iteration-0 baseline is measured once on val using the untrained template team under none and is shared by all SkillOpt runs with the same team composition. A fingerprint of the settings that affect the result is stored with it, and the baseline is re-measured if the fingerprint changes. After an accepted update, the baseline becomes the accepted score. For a newly added role, the baseline is set to its success rate on the tasks where it was called during L3 gate scoring.

\subsubsection{L3 Gate (single\_check)}
The L3 gate first merges the proposals from all sessions of the iteration. For the same file, the last proposal wins, and the config and prompt of a new role are bundled into a single role addition. It then drops invalid proposals: removing the chairman, a new role without a config, and invalid paths. Under competition, proposals that change the chairman are also dropped, since the ranking decides the chairman.

Next, the decision rule of the team's structure of interdependence (Table~\ref{tab:social}) determines which proposals to keep and in what order. The decision is made by the actual model of the team's chairman, and if its response cannot be parsed, all proposals are kept. Following this order, at most two role changes and two patch bundles are kept, so that each gate judges only a few changes; this cap applies to every condition, including none. In preliminary runs without the cap, teams that started with six roles grew to as many as 17 (LoCoBench, Meta-Team weakstrong).

The current team and the team with all remaining changes applied then each solve the same 20 val problems with the same seeds (40 sessions, with a separate working directory for each team). With $s_b$ successes out of $n_b$ for the current team and $s_c$ out of $n_c$ for the modified team, let $b = s_b/n_b$ and $X \sim \mathrm{Bin}(n_c, b)$. All changes are reverted $\iff s_c < b\,n_c \;\wedge\; P(X \le s_c) \le \alpha$; otherwise, all are applied. If fewer than 10 sessions complete without crashing, nothing is applied. The team folder is backed up before changes are applied and restored if applying them fails.

\section{Structure of Interdependence Specification}\label{app:interdependence}
\begin{table*}[t]
\centering
\small
\resizebox{\textwidth}{!}{%
\begin{tabular}{lllll}
\toprule
 & None & Collaboration & Conformity & Competition \\
\midrule
L2 sharing & Disabled & Active & Only when blocked & Minimal \\
L2 note commit & -- & Stage $\rightarrow$ challenge $\rightarrow$ commit & Direct write & Stage $\rightarrow$ challenge $\rightarrow$ commit \\
Comparative scores shown & -- & -- & -- & Yes \\
Tier 1 & All roles & All roles & Chairman & Top scorer \\
L3 proposal right & All roles & All roles & Chairman & Tier 1 \\
L3 challenge tool & No & Yes & No & Yes \\
Peer review & No & Majority, 180\,s & No & Majority, 180\,s \\
End-of-step decision & Keep all & Drop if OPPOSE $>$ SUPPORT & Chairman alone & Chairman alone \\
Order under the cap & Arrival & SUPPORT$-$OPPOSE & Chairman's priority & Chairman's priority \\
Chairman change & No & No & No & Top scorer, every step \\
\bottomrule
\end{tabular}%
}
\caption{Full specification of the structure of interdependence. Scoring and revert rules are identical across conditions. Gate sessions run under the same condition, tiers, chairman, and constitution as training sessions.}
\label{tab:social}
\end{table*}

We confirmed over repeated calls that both the weak and strong models follow the response format of the decision prompt (\textit{\{"keep": [...], "order": [...]\}}). Under collaboration, the decision follows an explicit rule that classifies each review as SUPPORT, OPPOSE, or NEUTRAL and counts them, so the decision model's own opinion is not taken into account.

\subsection{Competition: Promotion and Chairman Change}\label{app:competition}
At the end of each iteration, a score is computed for each role, and the sole top scorer becomes Tier 1 and the chairman for the next iteration; in case of a tie, the current Tier 1 stays. The score depends on the mechanism. Under SkillOpt, it is the current gate score of the role's skill (the iteration-0 baseline or the last accepted score); under MemRL, it is the mean $Q$ of the role's memory bank; and under Meta-Team, it is the success rate on tasks the role took part in during its single training iteration.

When the chairman changes, the original chairman is recorded in the team configuration, and a notice of the change is added to the session prompts. The new chairman is told that it has taken over the chairman's authority (receiving the task, calling roles, and submitting the final answer) and that it can still call the original chairman as a teammate. The original chairman and the other roles are told who the chairman is and that their own duties are unchanged. In sessions without a promotion, the prompts are identical byte for byte. The promoted chairman decides on its own whom to call, and we treat this behavior itself as an object of observation.

The score has no minimum sample size, so a rarely called role can reach first place through an extreme value from a small sample (e.g., 1/1). We keep this rule and report, for every promotion, the role, its score, the sample size, and performance before and after (Table~\ref{tab:promotions}). Across the three benchmarks, the chairman changed 20 times in 16 of the 27 competition runs. Of the 12 changes under SkillOpt and Meta-Team, whose scores count tasks, five rested on three or fewer tasks, and on GAIA, file\_agent, which takes part in only 2--9\% of sessions (Table~\ref{tab:workload}), became the chairman three times. The new chairman ran on the weak model in 13 of the 20 changes, including every change in OfficeQA Pro weakstrong, where the strong roles are the readers.

\providecommand{\tbd}{\textcolor{gray}{\textit{tbd}}}
\begin{table*}[t]
\centering
\setlength{\tabcolsep}{4pt}
\resizebox{\textwidth}{!}{%
\begin{tabular}{lllcl c cc c c c}
\toprule
 & & & & & & \multicolumn{2}{c}{New chairman} & Runner-up & Batch success (\%) & avg@3 \\
\cmidrule(lr){7-8}
Bench. & Mechanism & Team & Step & Chairman change & Model & score & $n$ & score ($n$) & before $\rightarrow$ after & (\%) \\
\midrule
GAIA & SkillOpt & onlyweak & 1 & plan\_agent $\rightarrow$ verifier\_agent$^\dagger$ & W & 0.67 & 3 & 0.20 (20) & 11.8 $\rightarrow$ 32.4 & 28.3 \\
GAIA & SkillOpt & weakstrong & 0 & plan\_agent $\rightarrow$ web\_agent & \textbf{S} & 0.56 & 16 & 0.50 (20) & 58.8 $\rightarrow$ 88.2 & \textbf{74.2} \\
GAIA & SkillOpt & onlystrong & 0 & plan\_agent $\rightarrow$ file\_agent & S & 1.00 & 1 & 0.75 (20) & 64.7 $\rightarrow$ 47.1 & 51.7 \\
GAIA & MemRL & onlyweak & 2 & plan\_agent $\rightarrow$ answer\_agent & W & $-0.218$ & 48 & $-0.220$ (49) & 23.5 $\rightarrow$ 41.2 & \multirow{2}{*}{13.3} \\
GAIA & MemRL & onlyweak & 3 & answer\_agent $\rightarrow$ file\_agent & W & 0.023 & 7 & $-0.203$ (63) & 41.2 $\rightarrow$ -- & \\
GAIA & Meta-Team & weakstrong & 0 & plan\_agent $\rightarrow$ web\_agent & \textbf{S} & 0.69 & 16 & 0.65 (17) & 64.7 $\rightarrow$ -- & \textbf{63.3} \\
GAIA & Meta-Team & onlystrong & 0 & plan\_agent $\rightarrow$ file\_agent & S & 1.00 & 1 & 0.71 (17) & 70.6 $\rightarrow$ -- & 43.3 \\
\midrule
LoCoBench & SkillOpt & onlyweak & 1 & planner $\rightarrow$ verifier$^\dagger$ & W & 1.00 & 1 & 0.70 (20) & 70.0 $\rightarrow$ 35.0 & 28.3 \\
LoCoBench & MemRL & onlystrong & 1 & planner $\rightarrow$ developer & S & 0.101 & 17 & 0.022 (33) & 45.0 $\rightarrow$ 77.5 & 83.5 \\
LoCoBench & Meta-Team & onlyweak & 0 & planner $\rightarrow$ reader-1 & W & 0.53 & 19 & 0.50 (20) & 50.0 $\rightarrow$ -- & 22.6 \\
\midrule
OfficeQA Pro & SkillOpt & onlyweak & 0 & planner $\rightarrow$ answer\_agent & W & 0.14 & 7 & 0.10 (10) & 10.0 $\rightarrow$ 8.3 & 2.2 \\
OfficeQA Pro & SkillOpt & weakstrong & 0 & planner $\rightarrow$ calculator & \textbf{W} & 0.38 & 13 & 0.35 (20) & 15.0 $\rightarrow$ 22.4 & \multirow{2}{*}{15.1} \\
OfficeQA Pro & SkillOpt & weakstrong & 1 & calculator $\rightarrow$ answer\_agent & \textbf{W} & 1.00 & 2 & 0.38 (13) & 20.0 $\rightarrow$ 23.6 & \\
OfficeQA Pro & SkillOpt & onlystrong & 1 & planner $\rightarrow$ calculator & S & 0.62 & 13 & 0.50 (2) & 50.0 $\rightarrow$ 35.7 & 33.9 \\
OfficeQA Pro & MemRL & onlyweak & 1 & planner $\rightarrow$ reader-2 & W & 0.000 & 9 & $-0.027$ (11) & 0.0 $\rightarrow$ 7.0 & \multirow{2}{*}{6.5} \\
OfficeQA Pro & MemRL & onlyweak & 3 & reader-2 $\rightarrow$ reader-1 & W & $-0.250$ & 15 & $-0.266$ (15) & 5.0 $\rightarrow$ -- & \\
OfficeQA Pro & MemRL & weakstrong & 1 & planner $\rightarrow$ answer\_agent & \textbf{W} & 0.080 & 25 & 0.077 (20) & 35.0 $\rightarrow$ 19.5 & \multirow{2}{*}{10.2} \\
OfficeQA Pro & MemRL & weakstrong & 2 & answer\_agent $\rightarrow$ verifier$^\dagger$ & \textbf{W} & 0.000 & 11 & $-0.027$ (31) & 9.1 $\rightarrow$ 30.0 & \\
OfficeQA Pro & MemRL & onlystrong & 3 & planner $\rightarrow$ calculator & S & 0.039 & 54 & 0.012 (57) & 40.0 $\rightarrow$ -- & 35.5 \\
OfficeQA Pro & Meta-Team & weakstrong & 0 & planner $\rightarrow$ calculator & \textbf{W} & 0.14 & 14 & 0.13 (15) & 10.0 $\rightarrow$ -- & 14.5 \\
\bottomrule
\end{tabular}%
}
\caption{\textbf{Chairman changes under competition.} \emph{Model}: the model of the new chairman (W: weak, S: strong). The score that decides the ranking is the skill's gate score for SkillOpt ($n$: val tasks), the mean $Q$ of the memory bank for MemRL ($n$: memories), and the success rate on the role's batch tasks for Meta-Team ($n$: tasks). Batch success is measured in the step of the change and averaged over the later steps; -- marks a change at the last step. LoCoBench and OfficeQA Pro batches differ across steps, so their values are not directly comparable. In weakstrong teams, the change helps when the strong role takes the chair (GAIA: 74.2 and 63.3, the best structure) and hurts when a weak role does (OfficeQA Pro: 15.1, 10.2, and 14.5, at or near the bottom). No other competition run changed its chairman. $^\dagger$Role added by L3 in an earlier step.}
\label{tab:promotions}
\end{table*}

\section{Execution Engine Details}\label{app:engine}
This section describes how sessions and reflection are run. 

\subsection{Session Execution}\label{app:session}
Up to six sessions run concurrently in training batches, gate scoring, and final evaluation. Ordinary messages are delivered when the recipient is waiting, whereas a phase-transition notice, if unread, is attached to the agent's next tool result and delivered immediately. If an agent sends the same recipient a message identical to its previous one, or retries a send that just failed (to a nonexistent recipient, to itself, or without content), the message is not sent and the agent waits until a new message arrives. Without this safeguard, agents never waited; they repeated the same message hundreds to thousands of times without ever receiving a reply.

The remaining steps, team messages, and time are given as the last message of each call and are not stored in the conversation history, and the system prompt stays fixed throughout a session to keep the prefix cache. Once the history exceeds 151 messages, only the latest 120 are kept. Only when the next call would exceed the model limit (200k minus the role's output allowance minus an 8k margin) are older tool outputs shortened, step by step; the system prompt, the task, and the agent's own messages are never truncated. Sessions terminated by the time or message limit are not graded and count as failures, since their working directories may be left half-written; a session in which agents stop on their own is graded on its final output (the chairman's last answer, or the files in solution/ on LoCoBench).


\subsection{Deferred Reflection}\label{app:reflection}
Reflection is deferred: sessions in a batch wait after finishing their task, and reflection starts once every task in the batch is done, three sessions at a time. Reflection has its own time budget, separate from task time: 300 seconds for L1, 900 for L2, and 900 for L3 (at most 2,160 seconds per session in total, including a margin). Debates are capped at 20 rounds in L2 and 30 in L3. Meta-Team is excluded, since its batch runs sequentially and it reflects right after each task.

We defer reflection because, when it ran right after each task, finished sessions entering L2 re-prefilled their whole history while other sessions were still solving tasks. The resulting prefill load raised timeouts even in gate sessions (48/120 vs.\ 21/120 at step 0 in one LoCoBench run).


\subsection{Seeds and Repeated Evaluation}\label{app:seeds}
Training and gate sessions are seeded by task ID, and final evaluation by a hash of (task ID, run), so all experiments share the same set of seeds. Since outcomes still vary across repeats (Appendix~\ref{app:infra}), we report avg@3 with its SD. The SD reflects evaluation noise only, not the variance of the single evolution run per condition. Going from three to five passes would shrink the standard error only modestly (from 2.9 to 2.25 points on the 99 LoCoBench test problems) at 67\% more cost.

\section{Infrastructure and Validation}\label{app:infra}
We serve the models on sets of two GPUs. In each set, one GPU hosts the weak model (memory fraction 0.90) and the bge-m3 embedding model (0.03), and the other hosts the strong model (0.90). Each of the four sets runs one experiment behind its own LiteLLM proxy. Calls to the strong model are about 1.1--1.3 times slower than calls to the weak model, measured by time to first token.

To validate the FP8 KV cache, we ran LoCoBench solo\_weak preevolve on the 99 test problems with BF16 and FP8 KV caches, using the same seeds and time limit. The success rate was 0.444 with BF16 and 0.545 with FP8 (paired McNemar $p = 0.11$), so we found no evidence of degradation. On the 67 problems with no timeout under either setting, the median session time dropped from 333 to 215 seconds, and the KV cache capacity of the weak model grew from about 298k to 578k tokens.

Even with seeds fixed per task, repeating the same experiment three times changed the outcome of 32 of the 99 LoCoBench test problems, likely due to numerical differences from GPU batch composition and the timing of asynchronous execution. This is why we report avg@3.
The main GAIA, LoCoBench, and OfficeQA experiments ran on machines with eight NVIDIA RTX PRO 6000 Blackwell GPUs (96\,GB), and the analysis experiments (Appendix~\ref{app:analysis_setup}) on eight RTX PRO 6000 Blackwell Workstation Edition GPUs (team composition and capability gap) and four RTX PRO 6000 Blackwell Server Edition GPUs (model family). We used vLLM 0.28.0 with FlashInfer 0.6.16.post3, Transformers 5.18.0, and PyTorch 2.13, and LiteLLM 1.101.0 as the proxy; the Mistral models ran on a separate install with vLLM 0.31.0. Each model server handles up to 16 concurrent sequences with a 200k-token context (131,072 tokens in the Gemma experiment). Thinking is turned off on every request through the chat template (\texttt{enable\_thinking=false}), which overrides any server default. The Mistral tokenizer does not accept this argument, so it is omitted for the Mistral models, whose reasoning is off by default (\texttt{reasoning\_effort=none}). The wall-clock time of one experiment on one GPU set was 4.5--16.0 hours for SkillOpt, 3.4--9.1 hours for MemRL, and 2.1--4.9 hours for Meta-Team on GAIA, and 14.5--37.2 hours for SkillOpt on LoCoBench.

\section{Patterns by Ability Band}
\label{app:patterns}
The results in Table~\ref{tab:main} follow how capable each agent is on the task. Before evolution, the weak model scores 2.7, 20.8, and 54.5 alone on OfficeQA Pro, GAIA, and LoCoBench, and the strong model 28.5, 51.7, and 73.1. Grouping agents into bands by this solo score (0--20, 20--50, 50--70, and 70+) reveals five main patterns in the best social scaffolding; the two remaining teams are discussed in Appendix~\ref{app:Identical5070}.
\begin{itemize}[leftmargin=*, topsep=0pt, itemsep=2pt, parsep=0pt]
    \item \textbf{P1. A near-zero member} (OfficeQA Pro onlyweak and weakstrong). No structure rescues the team; even the best mixed team (21.5) stays below the strong agent alone (28.5).
    \item \textbf{P2. Identical 20--50 agents} (GAIA onlyweak, OfficeQA Pro onlystrong). Structures help little, adding at most 8.3 points over evolution.
    \item \textbf{P3. 20--50 mixed with 50--70} (GAIA weakstrong). Competition helps most with SkillOpt and Meta-Team, lifting the team to 74.2 and 63.3, well above any other structure; with MemRL, where competition leaves the chair weak, the best structure adds only 8.3 points over evolution, no more than in P2.
    \item \textbf{P4. 50--70 mixed with 70+} (LoCoBench weakstrong). Already strong before evolution, the team gains 1.7--6.0 points over \emph{Pre} under collaboration and competition, while conformity ranks last among the three structures under every mechanism.
    \item \textbf{P5. Identical 70+ agents} (LoCoBench onlystrong). Every mechanism has a structure that helps strongly, but a different one: conformity with SkillOpt (87.2), collaboration with Meta-Team (72.4), and competition with MemRL (83.5).
\end{itemize}

\subsection{Teams of Identical 50–70 Agents}\label{app:Identical5070}

The two teams of identical 50–70 agents fit none of the five patterns above. Evolution alone already lifts GAIA onlystrong by 14.2–16.7 points, and no structure adds to it under Meta-Team or SkillOpt, whereas LoCoBench onlyweak gains most from collaboration under Meta-Team and SkillOpt (59.3 and 54.9) and collapses under competition (22.6 and 28.3).

\section{Process Metrics}\label{app:process_metrics}
Social scaffolding can change final performance only through what happens during evolution. We therefore trace its effect in stages: the structure of interdependence changes how agents reflect, reflection changes the team structure, and the team structure changes how agents collaborate and how efficiently they solve tasks, which in turn determines performance. We define metrics for each stage so that a difference in performance can be attributed to a specific change in process. All metrics are computed from three records kept for every experiment: per-session event logs, which record every message, tool call, LLM call with its token counts, timeout, and session duration; per-iteration snapshots of the team, including its roles, prompts, and iteration result; and the decisions of the L1 and L3 gates.

\subsection{Reflection}\label{app:pm_reflection}
The structure of interdependence acts most directly on reflection (Table~\ref{tab:pm_reflection}). Under none, L2 is skipped and L3 proposals are kept without discussion, so the metrics that depend on peer interaction are defined only for collaboration, conformity, and competition.

\begin{table}[htbp]
\centering
\large
\resizebox{\columnwidth}{!}{%
\begin{tabular}{p{4.2cm}p{9cm}}
\toprule
Metric & Definition \\
\midrule
Challenge rate & Challenges to a teammate's note or reflection, divided by the number of proposals \\
Withdrawal rate & Notes withdrawn by their author, divided by the number of notes \\
Reflection adoption rate & Reflections applied, divided by reflections proposed \\
Skip rate & Share of sessions in which the agent skips L2 or L3 reflection \\
Team-improvement suggestions & Suggestions sent to the chairman by Tier~2 agents \\
Teammate-model updates & Updates to teammate profiles and collaboration notes \\
\bottomrule
\end{tabular}%
}
\caption{Reflection metrics, counted per session and averaged per iteration.}
\label{tab:pm_reflection}
\end{table}

\subsection{Team Structure}\label{app:pm_structure}
Table~\ref{tab:pm_structure} lists the metrics for changes made through L3. Proposals are counted in units of one role addition, one role removal, one constitution patch, or one patch to an existing role, matching the units that the L3 gate merges and caps (Appendix~\ref{app:gate}). Initial team sizes differ across benchmarks (Table~\ref{tab:roles}), so we report team size both as an absolute number and relative to the initial team.

\begin{table}[htbp]
\centering
\large
\resizebox{\columnwidth}{!}{%
\begin{tabular}{p{4.2cm}p{9cm}}
\toprule
Metric & Definition \\
\midrule
Roles added and removed & Per iteration and cumulative, counting only changes that pass the L3 gate \\
Final team size & Number of roles after the last iteration \\
Proposals and adoption rate & Proposals made, proposals kept by the decision rule, and proposals applied after the gate \\
Change types & Shares of role additions, constitution patches, and patches to existing roles \\
Gate margin & Validation successes of the current and modified teams and the $p$-value of each L3 gate \\
Usage of added roles & Share of later sessions in which an added role is actually called \\
\bottomrule
\end{tabular}%
}
\caption{Team-structure metrics for L3.}
\label{tab:pm_structure}
\end{table}

\subsection{Role-level Updates}\label{app:pm_l1}
L1 updates are recorded per role and per iteration: for SkillOpt, the number of candidate skills accepted by the L1 gate and each role's gate score over iterations; for MemRL, the size of each role's memory bank, its mean $Q$, and the number of value updates. Under competition, we also record each chairman change, its step, and the score margin (Table~\ref{tab:promotions}).

\subsection{Collaboration}\label{app:pm_collaboration}
Collaboration metrics describe how a team solves a single task (Table~\ref{tab:pm_collaboration}). Under competition, the chairman can change across iterations, so chairman-mediated messages are counted with respect to the chairman of that session.

\begin{table}[htbp]
\centering
\large
\resizebox{\columnwidth}{!}{%
\begin{tabular}{p{3.6cm}p{6.4cm}p{4cm}}
\toprule
Metric & Definition & What it indicates \\
\midrule
Messages & Messages sent per session & Volume of communication \\
Active roles & Roles called per session & Team size in use \\
Speaker concentration & Gini coefficient of each role's share of messages & Hierarchical vs.\ flat \\
Chairman share & Share of messages between the chairman and a teammate, among all messages & Hub vs.\ mesh \\
Coordination overhead & Calls to wait for replies or check teammates' status, divided by calls to work tools (file, shell, web search) & Waiting relative to working \\
Communication failures & Sends blocked as repeats and sends to roles that do not exist & Confusion as the team changes \\
\bottomrule
\end{tabular}%
}
\caption{Collaboration metrics, computed per session.}
\label{tab:pm_collaboration}
\end{table}

\subsection{Efficiency and Failures}\label{app:pm_efficiency}
Table~\ref{tab:pm_efficiency} lists the efficiency and failure metrics. Timeouts and session durations are reported separately for training and final evaluation.

\begin{table}[htbp]
\centering
\large
\resizebox{\columnwidth}{!}{%
\begin{tabular}{p{4.2cm}p{9cm}}
\toprule
Metric & Definition \\
\midrule
Batch success rate & Success rate of each training batch \\
Timeout rate & Sessions that hit the 900-second limit, divided by all sessions \\
Session duration & Median and 90th percentile of session time \\
Cost per session & Messages, LLM calls, and input and output tokens \\
Abnormal endings & Sessions in which the chairman exits without submitting an answer, and sessions whose last message is used as a fallback answer \\
Context trimming & Times per session that older tool outputs are shortened (Appendix~\ref{app:engine}) \\
\bottomrule
\end{tabular}%
}
\caption{Efficiency and failure metrics.}
\label{tab:pm_efficiency}
\end{table}

\begin{table*}[t]
\centering
\small
\setlength{\tabcolsep}{3pt}
\resizebox{\textwidth}{!}{%
\begin{tabular}{ll cc c cccc c cccc c cccc}
\toprule
 & & \multicolumn{2}{c}{Solo} & & \multicolumn{4}{c}{onlyweak} & & \multicolumn{4}{c}{weakstrong} & & \multicolumn{4}{c}{onlystrong} \\
\cmidrule(lr){3-4} \cmidrule(lr){6-9} \cmidrule(lr){11-14} \cmidrule(lr){16-19}
Mechanism & Step & Weak & Strong & & None & Collab. & Conform. & Compet. & & None & Collab. & Conform. & Compet. & & None & Collab. & Conform. & Compet. \\
\midrule
\multicolumn{19}{l}{\textit{GAIA (17 problems per step, the same at every step)}} \\
\midrule
\multirow{4}{*}{SkillOpt} & S0 & 29.4 & 70.6 & & 5.9 & 35.3 & 11.8 & 17.6 & & 58.8 & 41.2 & 47.1 & 58.8 & & 58.8 & 70.6 & 82.4 & 64.7 \\
 & S1 & 35.3 & 58.8 & & 29.4 & 17.6 & 23.5 & 11.8 & & 41.2 & 41.2 & 58.8 & 94.1 & & 70.6 & 76.5 & 82.4 & 35.3 \\
 & S2 & 29.4 & 70.6 & & 29.4 & 23.5 & 23.5 & 23.5 & & 58.8 & 58.8 & 58.8 & 88.2 & & 70.6 & 76.5 & 70.6 & 47.1 \\
 & S3 & 41.2 & 70.6 & & 35.3 & 29.4 & 11.8 & 41.2 & & 52.9 & 70.6 & 52.9 & 82.4 & & 64.7 & 82.4 & 88.2 & 58.8 \\
\cmidrule(lr){1-19}
\multirow{4}{*}{MemRL} & S0 & 35.3 & 70.6 & & 11.8 & 17.6 & 29.4 & 11.8 & & 35.3 & 47.1 & 58.8 & 47.1 & & 64.7 & 52.9 & 52.9 & 52.9 \\
 & S1 & 41.2 & 70.6 & & 41.2 & 29.4 & 11.8 & 29.4 & & 58.8 & 70.6 & 52.9 & 76.5 & & 76.5 & 82.4 & 76.5 & 64.7 \\
 & S2 & 35.3 & 82.4 & & 17.6 & 35.3 & 23.5 & 23.5 & & 58.8 & 64.7 & 52.9 & 76.5 & & 70.6 & 76.5 & 70.6 & 70.6 \\
 & S3 & 47.1 & 70.6 & & 41.2 & 41.2 & 17.6 & 41.2 & & 64.7 & 76.5 & 58.8 & 70.6 & & 82.4 & 41.2 & 76.5 & 70.6 \\
\cmidrule(lr){1-19}
\multirow{1}{*}{Meta-Team} & S0 & 23.5 & 70.6 & & 11.8 & 23.5 & 17.6 & 29.4 & & 52.9 & 52.9 & 52.9 & 64.7 & & 52.9 & 64.7 & 58.8 & 70.6 \\
\midrule[\heavyrulewidth]
\multicolumn{19}{l}{\textit{LoCoBench (20 problems per step)}} \\
\midrule
\multirow{4}{*}{SkillOpt} & S0 & 75.0 & 80.0 & & 40.0 & 50.0 & 60.0 & 45.0 & & 85.0 & 75.0 & 85.0 & 85.0 & & 50.0 & 55.0 & 55.0 & 65.0 \\
 & S1 & 60.0 & 85.0 & & 40.0 & 70.0 & 50.0 & 70.0 & & 70.0 & 80.0 & 80.0 & 75.0 & & 65.0 & 75.0 & 70.0 & 60.0 \\
 & S2 & 55.0 & 90.0 & & 35.0 & 40.0 & 60.0 & 35.0 & & 65.0 & 95.0 & 90.0 & 95.0 & & 75.0 & 75.0 & 90.0 & 90.0 \\
 & S3 & 45.0 & 80.0 & & 45.0 & 50.0 & 40.0 & 35.0 & & 80.0 & 90.0 & 85.0 & 80.0 & & 70.0 & 90.0 & 85.0 & 75.0 \\
\cmidrule(lr){1-19}
\multirow{4}{*}{MemRL} & S0 & 55.0 & 70.0 & & 50.0 & 40.0 & 55.0 & 60.0 & & 70.0 & 70.0 & 90.0 & 85.0 & & 65.0 & 55.0 & 60.0 & 65.0 \\
 & S1 & 55.0 & 75.0 & & 25.0 & 50.0 & 50.0 & 60.0 & & 90.0 & 85.0 & 80.0 & 75.0 & & 60.0 & 55.0 & 60.0 & 45.0 \\
 & S2 & 60.0 & 70.0 & & 45.0 & 55.0 & 30.0 & 45.0 & & 80.0 & 95.0 & 90.0 & 90.0 & & 80.0 & 65.0 & 60.0 & 75.0 \\
 & S3 & 55.0 & 85.0 & & 35.0 & 45.0 & 30.0 & 50.0 & & 80.0 & 95.0 & 95.0 & 85.0 & & 55.0 & 75.0 & 60.0 & 80.0 \\
\cmidrule(lr){1-19}
\multirow{1}{*}{Meta-Team} & S0 & 65.0 & 90.0 & & 85.0 & 55.0 & 65.0 & 50.0 & & 75.0 & 85.0 & 85.0 & 85.0 & & 80.0 & 65.0 & 65.0 & 70.0 \\
\midrule[\heavyrulewidth]
\multicolumn{19}{l}{\textit{OfficeQA Pro (20 problems per step; 11 at S2)}} \\
\midrule
\multirow{4}{*}{SkillOpt} & S0 & 0.0 & 25.0 & & 5.0 & 5.0 & 5.0 & 10.0 & & 15.0 & 25.0 & 15.0 & 15.0 & & 30.0 & 25.0 & 35.0 & 40.0 \\
 & S1 & 10.0 & 40.0 & & 5.0 & 0.0 & 0.0 & 20.0 & & 30.0 & 25.0 & 15.0 & 20.0 & & 55.0 & 55.0 & 45.0 & 50.0 \\
 & S2 & 9.1 & 36.4 & & 0.0 & 0.0 & 9.1 & 0.0 & & 18.2 & 27.3 & 9.1 & 27.3 & & 54.5 & 63.6 & 45.5 & 36.4 \\
 & S3 & 5.0 & 35.0 & & 0.0 & 0.0 & 0.0 & 5.0 & & 25.0 & 30.0 & 15.0 & 20.0 & & 35.0 & 50.0 & 50.0 & 35.0 \\
\cmidrule(lr){1-19}
\multirow{4}{*}{MemRL} & S0 & 5.0 & 20.0 & & 5.0 & 5.0 & 0.0 & 5.0 & & 5.0 & 10.0 & 20.0 & 20.0 & & 40.0 & 30.0 & 20.0 & 35.0 \\
 & S1 & 0.0 & 65.0 & & 0.0 & 0.0 & 0.0 & 0.0 & & 30.0 & 35.0 & 25.0 & 35.0 & & 55.0 & 60.0 & 55.0 & 65.0 \\
 & S2 & 0.0 & 63.6 & & 0.0 & 0.0 & 9.1 & 9.1 & & 18.2 & 18.2 & 27.3 & 9.1 & & 63.6 & 63.6 & 63.6 & 54.5 \\
 & S3 & 0.0 & 35.0 & & 5.0 & 0.0 & 0.0 & 5.0 & & 15.0 & 20.0 & 30.0 & 30.0 & & 55.0 & 55.0 & 40.0 & 40.0 \\
\cmidrule(lr){1-19}
\multirow{1}{*}{Meta-Team} & S0 & 5.0 & 40.0 & & 0.0 & 0.0 & 5.0 & 0.0 & & 10.0 & 30.0 & 15.0 & 10.0 & & 25.0 & 40.0 & 35.0 & 30.0 \\
\bottomrule
\end{tabular}%
}
\caption{Training batch success rate (\%) at each step (S0--S3); Meta-Team runs a single step. On GAIA every step solves the same 17 train problems in a new order, so steps are comparable; on LoCoBench and OfficeQA Pro each step solves different problems, and step 2 of OfficeQA Pro has only 11 problems (Appendix~\ref{app:benchmarks}).}
\label{tab:batch_success}
\end{table*}

The batch success rate is not comparable across iterations, because each iteration solves different problems (except on GAIA, Appendix~\ref{app:benchmarks}) and the last batch of an epoch is smaller (11 problems at iteration 2 of OfficeQA). We use it only as a secondary signal.

\begin{table*}[t]
\centering
\small
\setlength{\tabcolsep}{3pt}
\resizebox{\textwidth}{!}{%
\begin{tabular}{ll cccc c cccc c cccc}
\toprule
 & & \multicolumn{4}{c}{GAIA} & & \multicolumn{4}{c}{LoCoBench} & & \multicolumn{4}{c}{OfficeQA Pro} \\
\cmidrule(lr){3-6} \cmidrule(lr){8-11} \cmidrule(lr){13-16}
Team & Structure & TO & Time & Calls & Msgs & & TO & Time & Calls & Msgs & & TO & Time & Calls & Msgs \\
\midrule
\multirow{5}{*}{onlyweak} & \emph{Pre} & 11.7 & 257 & 45.8 & 15.4 & & 17.8 & 530 & 106.6 & 26.9 & & 79.6 & 902 & 96.5 & 17.2 \\
 & None & 8.6 & 271 & 66.7 & 13.1 & & 23.4 & 581 & 112.6 & 27.0 & & 53.2 & 728 & 90.5 & 19.9 \\
 & Collab. & 5.0 & 226 & 60.3 & 9.1 & & 19.1 & 575 & 99.3 & 26.4 & & 46.8 & 773 & 79.1 & 14.8 \\
 & Conform. & 8.3 & 259 & 48.2 & 8.2 & & 26.1 & 619 & 127.4 & 33.5 & & 86.4 & 908 & 77.0 & 14.7 \\
 & Compet. & 5.3 & 168 & 75.0 & 18.7 & & 35.5 & 738 & 142.3 & 40.6 & & 41.7 & 692 & 106.8 & 27.3 \\
\midrule
\multirow{5}{*}{weakstrong} & \emph{Pre} & 7.5 & 132 & 46.8 & 20.4 & & 5.4 & 341 & 80.0 & 19.6 & & 44.6 & 772 & 109.8 & 15.1 \\
 & None & 3.1 & 66 & 41.6 & 6.6 & & 7.8 & 374 & 93.1 & 23.2 & & 36.0 & 634 & 110.7 & 19.1 \\
 & Collab. & 3.3 & 57 & 36.3 & 5.2 & & 4.6 & 327 & 81.9 & 19.6 & & 39.1 & 750 & 129.7 & 21.9 \\
 & Conform. & 1.7 & 139 & 55.7 & 8.3 & & 5.6 & 395 & 98.9 & 23.2 & & 42.6 & 712 & 113.7 & 17.8 \\
 & Compet. & 2.8 & 48 & 25.5 & 2.3 & & 3.9 & 342 & 78.7 & 18.0 & & 23.8 & 301 & 128.8 & 33.2 \\
\midrule
\multirow{5}{*}{onlystrong} & \emph{Pre} & 22.5 & 182 & 42.4 & 6.0 & & 23.2 & 534 & 73.3 & 4.8 & & 34.4 & 525 & 91.9 & 11.8 \\
 & None & 14.5 & 136 & 38.6 & 5.5 & & 22.7 & 545 & 56.8 & 3.7 & & 33.1 & 336 & 81.4 & 12.1 \\
 & Collab. & 15.0 & 150 & 43.2 & 5.7 & & 16.1 & 471 & 37.4 & 1.7 & & 25.4 & 355 & 88.9 & 12.4 \\
 & Conform. & 15.3 & 169 & 40.6 & 6.1 & & 14.7 & 447 & 43.4 & 2.1 & & 26.9 & 373 & 78.4 & 12.5 \\
 & Compet. & 20.8 & 159 & 45.1 & 5.1 & & 11.9 & 444 & 35.6 & 1.9 & & 30.3 & 129 & 51.3 & 7.1 \\
\bottomrule
\end{tabular}%
}
\caption{Efficiency in the final evaluation. \emph{TO}: sessions stopped by the 900-second limit (\%); \emph{Time}: median session time (s); \emph{Calls}: LLM calls per session; \emph{Msgs}: messages sent per session. Evolved rows average the Meta-Team, SkillOpt, and MemRL runs of each structure; \emph{Pre} is the same team before evolution.}
\label{tab:efficiency}
\end{table*}

Table~\ref{tab:batch_success} lists the batch success rate at each iteration and Table~\ref{tab:efficiency} the efficiency of the final evaluation. On GAIA, where every iteration solves the same problems, the batch success of SkillOpt weakstrong under competition jumps from 58.8 to 94.1 right after competition hands the chair to web\_agent (Table~\ref{tab:promotions}), whereas the other structures stay between 41.2 and 70.6. In the final evaluation, timeouts fall most where performance rises most: in LoCoBench onlystrong they drop from 23.2\% before evolution to 3.7--12.8\% under SkillOpt (Table~\ref{tab:process}), while the score rises from 63.0 to 76.8--87.2.

\subsection{Linking Process to Performance}\label{app:pm_linking}
We connect the process metrics to performance in three ways. First, across all evolved conditions of the three benchmarks, we correlate the change in avg@3 from the preevolve baseline with the change in team size, timeout rate, and messages per session. Second, we compare outcomes task by task before and after evolution, which is less affected by evaluation noise than comparing means (Appendix~\ref{app:seeds}). Third, for conditions with large effects, we compare the evolved skills and team configurations directly with the templates.
Where performance is near the floor, as for the OfficeQA onlyweak team, differences in avg@3 carry little information, and we describe these conditions through the process metrics instead.

\begin{table}[t]
\centering
\small
\setlength{\tabcolsep}{4pt}
\resizebox{\columnwidth}{!}{%
\begin{tabular}{lccccc}
\toprule
 & $n$ & $\Delta$Team size & $\Delta$TO & $\Delta$Messages & $\Delta$Calls \\
\midrule
GAIA & 36 & \textbf{$-$0.59} & \textbf{$-$0.46} & 0.06 & $-$0.28 \\
LoCoBench & 36 & \textbf{$-$0.45} & \textbf{$-$0.83} & \textbf{$-$0.65} & \textbf{$-$0.67} \\
OfficeQA Pro & 36 & $-$0.17 & $-$0.22 & 0.00 & $-$0.16 \\
\midrule
Pooled & 108 & \textbf{$-$0.32} & \textbf{$-$0.42} & \textbf{$-$0.36} & \textbf{$-$0.33} \\
\bottomrule
\end{tabular}%
}
\caption{Spearman correlation between the change in avg@3 from \emph{Pre} and the change in each process metric, over all evolved runs (36 per benchmark). \textbf{Bold}: $p<0.05$.}
\label{tab:linking_corr}
\end{table}

Across the 108 evolved runs, a larger gain over \emph{Pre} goes with a smaller team and fewer timeouts (Table~\ref{tab:linking_corr}; pooled Spearman $\rho = -0.32$ and $-0.42$). The link is strongest on LoCoBench, where the change in timeouts alone tracks the change in score ($\rho = -0.83$), and absent on OfficeQA Pro, where most teams stay near their floor. Messages per session follow performance on LoCoBench but not on GAIA, where the best runs replace the relay through the chairman with a single working agent.


\begin{table*}[t]
\centering
\small
\setlength{\tabcolsep}{4pt}
\resizebox{\textwidth}{!}{%
\begin{tabular}{lll l cc c cc}
\toprule
 & & & & \multicolumn{2}{c}{Best structure vs.\ none} & & \multicolumn{2}{c}{None vs.\ \emph{Pre}} \\
\cmidrule(lr){5-6} \cmidrule(lr){8-9}
Bench. & Mechanism & Team & Best structure & $+$/$-$ & $p$ & & $+$/$-$ & $p$ \\
\midrule
GAIA & Meta-Team & onlyweak & Compet. & +17/$-$10 & 0.248 & & +14/$-$14 & 1.000 \\
GAIA & Meta-Team & weakstrong & Compet. & +26/$-$12 & \textbf{0.034} & & +23/$-$19 & 0.644 \\
GAIA & Meta-Team & onlystrong & Collab. & +15/$-$18 & 0.728 & & +28/$-$8 & \textbf{0.001} \\
GAIA & SkillOpt & onlyweak & Compet. & +17/$-$7 & 0.064 & & +10/$-$12 & 0.832 \\
GAIA & SkillOpt & weakstrong & Compet. & +34/$-$7 & \textbf{<0.001} & & +21/$-$17 & 0.627 \\
GAIA & SkillOpt & onlystrong & Conform. & +13/$-$17 & 0.585 & & +25/$-$6 & \textbf{<0.001} \\
GAIA & MemRL & onlyweak & Collab. & +18/$-$11 & 0.265 & & +11/$-$15 & 0.557 \\
GAIA & MemRL & weakstrong & Conform. & +22/$-$12 & 0.121 & & +17/$-$15 & 0.860 \\
GAIA & MemRL & onlystrong & Collab. & +16/$-$14 & 0.856 & & +25/$-$8 & \textbf{0.005} \\
\midrule
LoCoBench & Meta-Team & onlyweak & Collab. & +93/$-$46 & \textbf{<0.001} & & +50/$-$78 & \textbf{0.017} \\
LoCoBench & Meta-Team & weakstrong & Compet. & +41/$-$25 & 0.064 & & +39/$-$37 & 0.909 \\
LoCoBench & Meta-Team & onlystrong & Collab. & +70/$-$29 & \textbf{<0.001} & & +40/$-$53 & 0.213 \\
LoCoBench & SkillOpt & onlyweak & Collab. & +78/$-$59 & 0.124 & & +59/$-$72 & 0.294 \\
LoCoBench & SkillOpt & weakstrong & Collab. & +59/$-$20 & \textbf{<0.001} & & +36/$-$59 & \textbf{0.023} \\
LoCoBench & SkillOpt & onlystrong & Conform. & +45/$-$14 & \textbf{<0.001} & & +69/$-$28 & \textbf{<0.001} \\
LoCoBench & MemRL & onlyweak & Compet. & +76/$-$57 & 0.118 & & +54/$-$82 & \textbf{0.020} \\
LoCoBench & MemRL & weakstrong & Compet. & +38/$-$25 & 0.130 & & +43/$-$43 & 1.000 \\
LoCoBench & MemRL & onlystrong & Compet. & +72/$-$15 & \textbf{<0.001} & & +46/$-$42 & 0.749 \\
\midrule
OfficeQA Pro & Meta-Team & onlyweak & Conform. & +2/$-$1 & 1.000 & & +0/$-$1 & 1.000 \\
OfficeQA Pro & Meta-Team & weakstrong & Conform. & +16/$-$15 & 1.000 & & +11/$-$8 & 0.648 \\
OfficeQA Pro & Meta-Team & onlystrong & Compet. & +25/$-$22 & 0.771 & & +16/$-$15 & 1.000 \\
OfficeQA Pro & SkillOpt & onlyweak & Compet. & +4/$-$1 & 0.375 & & +1/$-$2 & 1.000 \\
OfficeQA Pro & SkillOpt & weakstrong & Collab. & +15/$-$24 & 0.200 & & +20/$-$12 & 0.215 \\
OfficeQA Pro & SkillOpt & onlystrong & Conform. & +22/$-$20 & 0.878 & & +19/$-$16 & 0.736 \\
OfficeQA Pro & MemRL & onlyweak & Compet. & +11/$-$3 & 0.057 & & +4/$-$2 & 0.688 \\
OfficeQA Pro & MemRL & weakstrong & Collab. & +13/$-$12 & 1.000 & & +10/$-$15 & 0.424 \\
OfficeQA Pro & MemRL & onlystrong & Compet. & +28/$-$17 & 0.135 & & +12/$-$20 & 0.215 \\
\bottomrule
\end{tabular}%
}
\caption{Task-level paired comparison on the 120 (GAIA), 297 (LoCoBench), and 186 (OfficeQA Pro) (task, rollout) pairs of the test set, which share seeds across runs. $+$/$-$: pairs solved only by the first / only by the second run; $p$: exact McNemar test. \emph{Best structure}: the best of collaboration, conformity, and competition. \textbf{Bold}: $p<0.05$.}
\label{tab:linking_paired}
\end{table*}

Task by task (Table~\ref{tab:linking_paired}), the best structure solves more (task, rollout) pairs than evolution alone in 24 of the 27 settings, and the difference is significant ($p<0.05$, exact McNemar test) in 7 of them. These are exactly the settings where Appendix~\ref{app:patterns} reports that structure matters: P3 with SkillOpt and Meta-Team (competition), P4 with SkillOpt (collaboration), P5 with all three mechanisms, and LoCoBench onlyweak with Meta-Team (collaboration; Appendix~\ref{app:Identical5070}). None falls in P1 or P2, where we report that structures help little or not at all, and P3 with MemRL, whose best structure is conformity because competition leaves the chair weak, is not significant ($p=0.12$). Evolution alone differs significantly from \emph{Pre} in 7 settings: it helps the three GAIA onlystrong teams and LoCoBench SkillOpt onlystrong, and hurts LoCoBench Meta-Team and MemRL onlyweak and SkillOpt weakstrong. Because the best structure is chosen after the fact from three and the three rollouts of a task are not independent, these $p$-values are optimistic; we read them as a check on which effects stand out from evaluation noise rather than as confirmatory tests.

\section{Additional Experiment Setup}\label{app:analysis_setup}
The three additional experiments use the same harness, hyperparameters, splits, and seeds as the main experiments. The team-composition and capability-gap experiments run on GAIA, and the model-family experiment runs on GAIA with Gemma-4 and on LoCoBench with Mistral. Only the models or the assignment of strong roles change, as described below, and web access is not restricted, as in the main experiments.

\begin{table}[htbp]
\centering
\large
\resizebox{\columnwidth}{!}{%
\begin{tabular}{lcccc}
\toprule
Role & Sessions & Input tokens & Output tokens & Tool calls \\
\midrule
web\_agent & 78--80 & 50--78 & 28--52 & 33--54 \\
plan\_agent (chairman) & 100 & 15--32 & 39--58 & 30--42 \\
answer\_agent & 97--99 & 5--15 & 9--12 & 12--22 \\
file\_agent & 2--9 & 0--2 & 0--4 & 0--3 \\
\bottomrule
\end{tabular}%
}
\caption{Workload of each GAIA role (\%): share of sessions it joins and of the team's tokens and tool calls, given as the range over the three teams before evolution.}
\label{tab:workload}
\end{table}

\paragraph{Team composition.} The main weakstrong team makes only web\_agent strong (1 strong : 3 weak). To make more roles strong, we add strong roles in order of workload, measured on the scored test sessions of the untrained onlyweak, weakstrong, and onlystrong teams (Table~\ref{tab:workload}). web\_agent consumes the largest share of input tokens, plan\_agent takes part in every session and produces a large share of the output tokens, answer\_agent takes part in almost every session but does little work, and file\_agent takes part in only 2--9\% of sessions. The 2:2 team therefore makes web\_agent and plan\_agent strong, and the 3:1 team additionally makes answer\_agent strong; file\_agent stays weak in all teams. Because plan\_agent is the chairman, the L3 decision in these two teams is made by the strong model. Workload alone does not show which role gains most from a stronger model, so we also evaluate, without evolution, a team in which only plan\_agent is strong and a team in which only answer\_agent is strong, and compare them with the main weakstrong team. The 2:2 and 3:1 teams are trained with all three mechanisms under all four structures.

\paragraph{Capability gap.} The weak model is replaced by Qwen3.5-2B (thinking off, as for all models), and the strong model is unchanged. On its official model card, Qwen3.5-2B reports 55.3 on MMLU-Pro and 51.6 on GPQA (measured in thinking mode), so the gap to the strong model grows from 6.8 to 29.9 points on MMLU-Pro and from 15.6 to 34.4 points on GPQA. We evaluate solo\_weak and weakstrong without evolution and train weakstrong with SkillOpt, Meta-Team, and MemRL under all four structures.

\paragraph{Model family.} The weak and strong models are replaced by Gemma-4 E4B and Gemma-4 26B-A4B. Their model cards report 69.4 and 82.6 on MMLU-Pro and 58.6 and 82.3 on GPQA Diamond (measured in thinking mode), gaps of 13.2 and 23.7 points. The two models are served under the same names as in the main experiments, so every stage fixed to the strong model (Table~\ref{tab:model_use}) uses Gemma-4 26B-A4B. Since E4B supports a 128K context, both models are served with a 131,072-token limit instead of 200k; in the main GAIA runs, only 0.57\% of sessions grew long enough to be affected by this limit. This experiment uses SkillOpt only, with solo\_weak and solo\_strong under none and onlyweak, weakstrong, and onlystrong under all four structures.

With thinking off, the Gemma-4 chat template starts each model turn with an empty thought block, and Gemma-4 26B-A4B often keeps emitting further empty thought blocks (\textit{<|channel>thought<channel|>}) until it reaches the output limit, without calling a tool. This happened mostly to the chairman: without evolution, 78 of the 120 test sessions of the onlystrong team hit the 900-second limit, and the team scored 20.0. Removing the empty block from the template did not stop the loop. We therefore suppress the token that opens a thought block (\textit{<|channel>}) for the strong model with a logit bias of $-100$, which keeps thinking off; with this, the same team scored 58.3. The weak model showed no such loop and is left unchanged. All reported Gemma runs that involve the strong model use this suppression.

On LoCoBench, the weak and strong models are replaced by Ministral-3-8B and Mistral-Small-4-119B-A6B (NVFP4). Their model cards report no benchmark on which both were evaluated in the same setting, so we characterize the pair by its solo scores on LoCoBench (58.2 and 77.4), which fall in the same 50–70 and 70+ bands as the Qwen pair. As with Gemma-4, every stage fixed to the strong model uses Mistral-Small-4, and the experiment uses SkillOpt only, with solo\_weak and solo\_strong under none and onlyweak, weakstrong, and onlystrong under all four structures.


\section{Additional Experiment Results}\label{app:analysis_results}
This section reports the three analysis experiments whose setup is given in Appendix~\ref{app:analysis_setup}. Each table reads its columns against the same team's \emph{Pre}, as in Table~\ref{tab:main}. Appendix~\ref{app:Identical5070} additionally discusses two teams from the main results. The results support the reading of Section~\ref{sec:analysis}: a structure of interdependence helps when it routes the work to a capable role, and which role is strong matters more than how many roles are strong.

\subsection{Team Composition}\label{app:res_comp}

\providecommand{\tbd}{\textcolor{gray}{\textit{tbd}}}
\providecommand{\na}{\textcolor{gray}{--}}
\begin{table}[!htbp]
\centering
\small
\resizebox{\columnwidth}{!}{%
\begin{tabular}{lc}
\toprule
Strong role & Pre \\
\midrule
web\_agent (main weakstrong) & 48.3\,{\scriptsize$\pm$1.4} \\
plan\_agent & 28.3\,{\scriptsize$\pm$2.9} \\
answer\_agent & 25.0\,{\scriptsize$\pm$2.5} \\
\midrule
web + plan (2:2) & 52.5\,{\scriptsize$\pm$6.6} \\
web + plan + answer (3:1) & 51.7\,{\scriptsize$\pm$3.8} \\
\bottomrule
\end{tabular}%
}
\caption{Which role is strong matters (GAIA, no evolution, avg@3, \%; mean\,$\pm$\,SD over three rollouts). Each row makes the listed role(s) strong and the rest weak.}
\label{tab:role}
\end{table}

Before evolution, the choice of the strong role matters far more than the number of strong roles (Table~\ref{tab:role}). Making only web\_agent strong gives 48.3, whereas making only plan\_agent strong gives 28.3 and making only answer\_agent strong gives 25.0, close to the onlyweak team (21.7). Adding strong roles on top of web\_agent barely helps: the 2:2 team scores 52.5 and the 3:1 team 51.7.

After evolution, the structure that helps most shifts once the chairman is strong (Table~\ref{tab:comp}). In the 1:3 team, where the weak plan\_agent chairs, competition is the best structure for SkillOpt and Meta-Team (74.2 and 63.3) because it moves the chair to the strong web\_agent (Section~\ref{sec:who}). In the 2:2 and 3:1 teams, plan\_agent is already strong, so competition has no weak chair to replace. With SkillOpt it moves the chair from one strong role to another (plan\_agent to web\_agent) and gains little (54.2 and 55.0); with Meta-Team it moves the chair to the weak file\_agent and gives the lowest score of its column (44.2 and 49.2); with MemRL the chair does not change. Instead, collaboration becomes the best structure for Meta-Team (60.8 and 63.3), for SkillOpt (57.5 and 56.7), and for MemRL in the 2:2 team (61.7). More strong roles do not raise the ceiling: the best 2:2 and 3:1 teams (61.7 and 63.3) stay below the 1:3 team under SkillOpt competition (74.2).

\providecommand{\tbd}{\textcolor{gray}{\textit{tbd}}}
\providecommand{\na}{\textcolor{gray}{--}}
\begin{table*}[t]
\centering
\setlength{\tabcolsep}{5pt}
\resizebox{\textwidth}{!}{%
\begin{tabular}{l c|ccc c|ccc c|ccc}
\toprule
& \multicolumn{4}{c}{1:3 main (strong: web)} & \multicolumn{4}{c}{2:2 (strong: web + plan)} & \multicolumn{4}{c}{3:1 (strong: web + plan + answer)} \\
\cmidrule(lr){2-5} \cmidrule(lr){6-9} \cmidrule(lr){10-13}
Structure & Pre & Meta-Team & SkillOpt & MemRL & Pre & Meta-Team & SkillOpt & MemRL & Pre & Meta-Team & SkillOpt & MemRL \\
\midrule
None & \multirow{4}{*}{48.3\,{\scriptsize$\pm$1.4}} & \cellcolor[RGB]{241,213,206}51.7\,{\scriptsize$\pm$2.9} & \cellcolor[RGB]{241,213,206}51.7\,{\scriptsize$\pm$2.9} & \cellcolor[RGB]{246,229,221}50.0\,{\scriptsize$\pm$4.3} & \multirow{4}{*}{52.5\,{\scriptsize$\pm$6.6}} & \cellcolor[RGB]{238,205,199}56.7\,{\scriptsize$\pm$1.4} & \cellcolor[RGB]{216,222,225}49.2\,{\scriptsize$\pm$5.2} & \cellcolor[RGB]{236,197,192}57.5\,{\scriptsize$\pm$2.5} & \multirow{4}{*}{51.7\,{\scriptsize$\pm$3.8}} & \cellcolor[RGB]{231,181,178}58.3\,{\scriptsize$\pm$2.9} & \cellcolor[RGB]{238,206,200}55.8\,{\scriptsize$\pm$8.0} & \cellcolor[RGB]{236,197,192}56.7\,{\scriptsize$\pm$7.6} \\
Collab. &  & \cellcolor[RGB]{198,210,219}43.3\,{\scriptsize$\pm$5.2} & \cellcolor[RGB]{214,128,130}62.5\,{\scriptsize$\pm$4.3} & \cellcolor[RGB]{243,221,214}50.8\,{\scriptsize$\pm$1.4} &  & \cellcolor[RGB]{225,164,163}\textbf{60.8}\,{\scriptsize$\pm$3.8} & \cellcolor[RGB]{236,197,192}\textbf{57.5}\,{\scriptsize$\pm$2.5} & \cellcolor[RGB]{223,156,155}\textbf{61.7}\,{\scriptsize$\pm$3.8} &  & \cellcolor[RGB]{215,132,134}\textbf{63.3}\,{\scriptsize$\pm$7.6} & \cellcolor[RGB]{236,197,192}\textbf{56.7}\,{\scriptsize$\pm$3.8} & \cellcolor[RGB]{243,221,214}54.2\,{\scriptsize$\pm$5.8} \\
Conform. &  & \cellcolor[RGB]{243,240,233}47.5\,{\scriptsize$\pm$0.0} & \cellcolor[RGB]{243,221,214}50.8\,{\scriptsize$\pm$3.8} & \cellcolor[RGB]{220,148,148}\textbf{58.3}\,{\scriptsize$\pm$5.8} &  & \cellcolor[RGB]{233,189,185}58.3\,{\scriptsize$\pm$1.4} & \cellcolor[RGB]{243,221,214}55.0\,{\scriptsize$\pm$6.6} & \cellcolor[RGB]{233,189,185}58.3\,{\scriptsize$\pm$5.2} &  & \cellcolor[RGB]{243,221,214}54.2\,{\scriptsize$\pm$3.8} & \cellcolor[RGB]{215,222,225}48.3\,{\scriptsize$\pm$2.9} & \cellcolor[RGB]{241,214,207}55.0\,{\scriptsize$\pm$2.5} \\
Compet. &  & \cellcolor[RGB]{214,128,130}\textbf{63.3}\,{\scriptsize$\pm$7.6} & \cellcolor[RGB]{214,128,130}\textbf{74.2}\,{\scriptsize$\pm$3.8} & \cellcolor[RGB]{243,221,214}50.8\,{\scriptsize$\pm$5.8} &  & \cellcolor[RGB]{164,187,208}44.2\,{\scriptsize$\pm$1.4} & \cellcolor[RGB]{246,229,221}54.2\,{\scriptsize$\pm$7.2} & \cellcolor[RGB]{236,197,192}57.5\,{\scriptsize$\pm$2.5} &  & \cellcolor[RGB]{225,228,228}49.2\,{\scriptsize$\pm$5.2} & \cellcolor[RGB]{241,214,207}55.0\,{\scriptsize$\pm$2.5} & \cellcolor[RGB]{225,164,163}\textbf{60.0}\,{\scriptsize$\pm$6.6} \\
\bottomrule
\end{tabular}%
}
\caption{Team composition on GAIA (avg@3, \%, mean\,$\pm$\,SD over three rollouts): the ratio of strong to weak roles in the weakstrong team. Strong roles are added in order of measured workload (web $>$ plan $>$ answer). Colors show the change from \emph{Pre} of the same team as in Table~\ref{tab:main}; \textbf{bold} marks the best structure of interdependence in each column.}
\label{tab:comp}
\end{table*}

\providecommand{\tbd}{\textcolor{gray}{\textit{tbd}}}
\providecommand{\na}{\textcolor{gray}{--}}
\begin{table*}[t]
\centering
\small
\setlength{\tabcolsep}{5pt}
\resizebox{\textwidth}{!}{%
\begin{tabular}{ll c|ccc c|ccc}
\toprule
& & \multicolumn{4}{c}{Main gap (weak = Qwen3-30B-A3B)} & \multicolumn{4}{c}{Larger gap (weak = Qwen3.5-2B)} \\
\cmidrule(lr){3-6} \cmidrule(lr){7-10}
Team & Structure & Pre & Meta-Team & SkillOpt & MemRL & Pre & Meta-Team & SkillOpt & MemRL \\
\midrule
solo\_weak & None & 20.8\,{\scriptsize$\pm$1.4} & \cellcolor[RGB]{251,246,236}20.8\,{\scriptsize$\pm$5.2} & \cellcolor[RGB]{251,246,236}20.8\,{\scriptsize$\pm$5.2} & \cellcolor[RGB]{243,240,233}20.0\,{\scriptsize$\pm$0.0} & 16.7\,{\scriptsize$\pm$5.8} & \cellcolor[RGB]{225,228,228}14.2\,{\scriptsize$\pm$5.8} & \cellcolor[RGB]{233,234,230}15.0\,{\scriptsize$\pm$6.6} & \cellcolor[RGB]{163,186,208}8.3\,{\scriptsize$\pm$3.8} \\
\midrule
\multirow{4}{*}{weakstrong} & None & \multirow{4}{*}{48.3\,{\scriptsize$\pm$1.4}} & \cellcolor[RGB]{241,213,206}51.7\,{\scriptsize$\pm$2.9} & \cellcolor[RGB]{241,213,206}51.7\,{\scriptsize$\pm$2.9} & \cellcolor[RGB]{246,229,221}50.0\,{\scriptsize$\pm$4.3} & \multirow{4}{*}{3.3\,{\scriptsize$\pm$3.8}} & \cellcolor[RGB]{248,237,228}\textbf{4.2}\,{\scriptsize$\pm$1.4} & \cellcolor[RGB]{246,229,221}5.0\,{\scriptsize$\pm$5.0} & \cellcolor[RGB]{243,221,214}5.8\,{\scriptsize$\pm$2.9} \\
 & Collab. &  & \cellcolor[RGB]{198,210,219}43.3\,{\scriptsize$\pm$5.2} & \cellcolor[RGB]{214,128,130}62.5\,{\scriptsize$\pm$4.3} & \cellcolor[RGB]{243,221,214}50.8\,{\scriptsize$\pm$1.4} &  & \cellcolor[RGB]{248,237,228}\textbf{4.2}\,{\scriptsize$\pm$3.8} & \cellcolor[RGB]{234,235,231}1.7\,{\scriptsize$\pm$2.9} & \cellcolor[RGB]{246,229,221}5.0\,{\scriptsize$\pm$2.5} \\
 & Conform. &  & \cellcolor[RGB]{243,240,233}47.5\,{\scriptsize$\pm$0.0} & \cellcolor[RGB]{243,221,214}50.8\,{\scriptsize$\pm$3.8} & \cellcolor[RGB]{220,148,148}\textbf{58.3}\,{\scriptsize$\pm$5.8} &  & \cellcolor[RGB]{251,246,236}3.3\,{\scriptsize$\pm$1.4} & \cellcolor[RGB]{243,221,214}\textbf{5.8}\,{\scriptsize$\pm$6.3} & \cellcolor[RGB]{234,235,231}1.7\,{\scriptsize$\pm$1.4} \\
 & Compet. &  & \cellcolor[RGB]{214,128,130}\textbf{63.3}\,{\scriptsize$\pm$7.6} & \cellcolor[RGB]{214,128,130}\textbf{74.2}\,{\scriptsize$\pm$3.8} & \cellcolor[RGB]{243,221,214}50.8\,{\scriptsize$\pm$5.8} &  & \cellcolor[RGB]{251,246,236}3.3\,{\scriptsize$\pm$1.4} & \cellcolor[RGB]{216,222,225}0.0\,{\scriptsize$\pm$0.0} & \cellcolor[RGB]{214,128,130}\textbf{65.8}\,{\scriptsize$\pm$6.3} \\
\bottomrule
\end{tabular}%
}
\caption{Larger weak--strong gap on GAIA (avg@3, \%, mean\,$\pm$\,SD over three rollouts). The strong model is Qwen3.6-35B-A3B in both settings; only the weak model changes. Colors show the change from \emph{Pre} of the same team as in Table~\ref{tab:main}; \textbf{bold} marks the best structure of interdependence in each column.}
\label{tab:gap}
\end{table*}

\subsection{Capability Gap}\label{app:res_gap}
As shown in Table~\ref{tab:gap}, with Qwen3.5-2B as the weak model, the solo weak agent scores 16.7, in the 0--20 band of P1, and evolving it alone does not help (14.2, 15.0, and 8.3 with Meta-Team, SkillOpt, and MemRL). The weakstrong team falls to 3.3 before evolution, far below both its own weak agent alone and the strong agent alone (51.7). The 2B chairman fails to lead: in 47 of the 120 test sessions the team hits the 900-second limit, and in 61 the chairman never submits an answer, repeatedly starting agents and sending messages without the required fields. Evolution leaves the team near zero (0.0--5.8) in 11 of the 12 settings, as P1 predicts. The exception is MemRL under competition, which hands the chair to the weak answer\_agent in the second iteration and to the strong web\_agent in the third; the training batch success then jumps from 17.0 to 82.0, web\_agent solves 118 of the 120 final sessions alone, and the team reaches 65.8, above even the strong agent alone. SkillOpt competition instead moves the chair only to answer\_agent, in the last iteration, and scores 0.0, and Meta-Team competition does not change the chair (3.3). The same team thus scores near zero under a weak chair and 65.8 once the strong role takes the chair, the extreme case of Section~\ref{sec:who}.

\subsection{Model Family}\label{app:res_family}
\providecommand{\tbd}{\textcolor{gray}{\textit{tbd}}}
\providecommand{\na}{\textcolor{gray}{--}}
\begin{table}[t]
\centering
\setlength{\tabcolsep}{5pt}
\small
\resizebox{\columnwidth}{!}{%
\begin{tabular}{ll c|c c|c}
\toprule
& & \multicolumn{2}{c}{Qwen (main)} & \multicolumn{2}{c}{Gemma-4} \\
\cmidrule(lr){3-4} \cmidrule(lr){5-6}
Team & Structure & Pre & SkillOpt & Pre & SkillOpt \\
\midrule
solo\_weak & None & 20.8\,{\scriptsize$\pm$1.4} & \cellcolor[RGB]{251,246,236}20.8\,{\scriptsize$\pm$5.2} & 27.5\,{\scriptsize$\pm$0.0} & \cellcolor[RGB]{249,238,229}28.3\,{\scriptsize$\pm$3.8} \\
solo\_strong & None & 51.7\,{\scriptsize$\pm$2.9} & \cellcolor[RGB]{225,228,228}49.2\,{\scriptsize$\pm$1.4} & 44.2\,{\scriptsize$\pm$5.2} & \cellcolor[RGB]{242,240,233}43.3\,{\scriptsize$\pm$3.8} \\
\midrule[\heavyrulewidth]
\multirow{4}{*}{onlyweak} & None & \multirow{4}{*}{21.7\,{\scriptsize$\pm$5.2}} & \cellcolor[RGB]{233,234,230}20.0\,{\scriptsize$\pm$4.3} & \multirow{4}{*}{20.0\,{\scriptsize$\pm$7.5}} & \cellcolor[RGB]{243,240,233}19.2\,{\scriptsize$\pm$3.8} \\
 & Collab. &  & \cellcolor[RGB]{225,228,228}19.2\,{\scriptsize$\pm$6.3} &  & \cellcolor[RGB]{249,238,229}20.8\,{\scriptsize$\pm$7.6} \\
 & Conform. &  & \cellcolor[RGB]{242,240,233}20.8\,{\scriptsize$\pm$1.4} &  & \cellcolor[RGB]{246,229,221}21.7\,{\scriptsize$\pm$6.3} \\
 & Compet. &  & \cellcolor[RGB]{231,181,178}\textbf{28.3}\,{\scriptsize$\pm$1.4} &  & \cellcolor[RGB]{220,148,148}\textbf{30.0}\,{\scriptsize$\pm$6.6} \\
\midrule
\multirow{4}{*}{weakstrong} & None & \multirow{4}{*}{48.3\,{\scriptsize$\pm$1.4}} & \cellcolor[RGB]{241,213,206}51.7\,{\scriptsize$\pm$2.9} & \multirow{4}{*}{35.8\,{\scriptsize$\pm$3.8}} & \cellcolor[RGB]{246,229,221}\textbf{37.5}\,{\scriptsize$\pm$0.0} \\
 & Collab. &  & \cellcolor[RGB]{214,128,130}62.5\,{\scriptsize$\pm$4.3} &  & \cellcolor[RGB]{199,210,219}30.8\,{\scriptsize$\pm$5.2} \\
 & Conform. &  & \cellcolor[RGB]{243,221,214}50.8\,{\scriptsize$\pm$3.8} &  & \cellcolor[RGB]{146,174,203}25.8\,{\scriptsize$\pm$8.0} \\
 & Compet. &  & \cellcolor[RGB]{214,128,130}\textbf{74.2}\,{\scriptsize$\pm$3.8} &  & \cellcolor[RGB]{216,222,225}32.5\,{\scriptsize$\pm$5.0} \\
\midrule
\multirow{4}{*}{onlystrong} & None & \multirow{4}{*}{45.0\,{\scriptsize$\pm$5.0}} & \cellcolor[RGB]{214,128,130}\textbf{60.8}\,{\scriptsize$\pm$5.8} & \multirow{4}{*}{58.3\,{\scriptsize$\pm$5.8}} & \cellcolor[RGB]{208,217,222}54.2\,{\scriptsize$\pm$3.8} \\
 & Collab. &  & \cellcolor[RGB]{230,180,177}51.7\,{\scriptsize$\pm$8.0} &  & \cellcolor[RGB]{234,235,231}56.7\,{\scriptsize$\pm$5.2} \\
 & Conform. &  & \cellcolor[RGB]{214,128,130}57.5\,{\scriptsize$\pm$6.6} &  & \cellcolor[RGB]{248,237,228}\textbf{59.2}\,{\scriptsize$\pm$10.1} \\
 & Compet. &  & \cellcolor[RGB]{230,180,177}51.7\,{\scriptsize$\pm$5.2} &  & \cellcolor[RGB]{164,187,208}50.0\,{\scriptsize$\pm$6.6} \\
\bottomrule
\end{tabular}%
}
\caption{Different model family on GAIA (avg@3, \%, mean\,$\pm$\,SD over three rollouts), SkillOpt only. Gemma-4: weak = E4B, strong = 26B-A4B; the SkillOpt optimizer is the family's strong model. Colors show the change from \emph{Pre} of the same team as in Table~\ref{tab:main}; \textbf{bold} marks the best structure of interdependence in each column.}
\label{tab:family}
\end{table}
\providecommand{\tbd}{\textcolor{gray}{\textit{tbd}}}
\providecommand{\na}{\textcolor{gray}{--}}
\begin{table}[t]
\centering
\setlength{\tabcolsep}{5pt}
\small
\resizebox{\columnwidth}{!}{%
\begin{tabular}{ll c|c c|c}
\toprule
& & \multicolumn{2}{c}{Qwen (main)} & \multicolumn{2}{c}{Mistral} \\
\cmidrule(lr){3-4} \cmidrule(lr){5-6}
Team & Structure & Pre & SkillOpt & Pre & SkillOpt \\
\midrule
solo\_weak & None & 54.5\,{\scriptsize$\pm$2.0} & \cellcolor[RGB]{198,210,219}49.5\,{\scriptsize$\pm$3.6} & 58.2\,{\scriptsize$\pm$1.5} & \cellcolor[RGB]{223,156,154}67.3\,{\scriptsize$\pm$3.5} \\
solo\_strong & None & 73.1\,{\scriptsize$\pm$1.5} & \cellcolor[RGB]{247,233,225}74.4\,{\scriptsize$\pm$2.5} & 77.4\,{\scriptsize$\pm$3.5} & \cellcolor[RGB]{242,216,209}80.5\,{\scriptsize$\pm$3.5} \\
\midrule[\heavyrulewidth]
\multirow{4}{*}{onlyweak} & None & \multirow{4}{*}{52.9\,{\scriptsize$\pm$3.5}} & \cellcolor[RGB]{205,214,221}48.5\,{\scriptsize$\pm$5.6} & \multirow{4}{*}{68.0\,{\scriptsize$\pm$3.2}} & \cellcolor[RGB]{241,213,206}\textbf{71.4}\,{\scriptsize$\pm$2.1} \\
 & Collab. &  & \cellcolor[RGB]{245,226,218}\textbf{54.9}\,{\scriptsize$\pm$6.1} &  & \cellcolor[RGB]{245,226,218}70.0\,{\scriptsize$\pm$3.1} \\
 & Conform. &  & \cellcolor[RGB]{176,195,212}45.8\,{\scriptsize$\pm$6.7} &  & \cellcolor[RGB]{205,214,221}63.6\,{\scriptsize$\pm$2.0} \\
 & Compet. &  & \cellcolor[RGB]{125,160,196}28.3\,{\scriptsize$\pm$1.7} &  & \cellcolor[RGB]{125,160,196}45.5\,{\scriptsize$\pm$3.6} \\
\midrule
\multirow{4}{*}{weakstrong} & None & \multirow{4}{*}{80.5\,{\scriptsize$\pm$1.5}} & \cellcolor[RGB]{169,190,210}72.7\,{\scriptsize$\pm$3.0} & \multirow{4}{*}{78.5\,{\scriptsize$\pm$2.9}} & \cellcolor[RGB]{235,196,191}83.5\,{\scriptsize$\pm$2.5} \\
 & Collab. &  & \cellcolor[RGB]{234,193,188}\textbf{85.9}\,{\scriptsize$\pm$5.6} &  & \cellcolor[RGB]{234,193,188}\textbf{83.8}\,{\scriptsize$\pm$3.0} \\
 & Conform. &  & \cellcolor[RGB]{236,236,231}79.1\,{\scriptsize$\pm$1.2} &  & \cellcolor[RGB]{236,200,194}83.2\,{\scriptsize$\pm$2.3} \\
 & Compet. &  & \cellcolor[RGB]{245,226,218}82.5\,{\scriptsize$\pm$4.6} &  & \cellcolor[RGB]{125,160,196}61.6\,{\scriptsize$\pm$4.4} \\
\midrule
\multirow{4}{*}{onlystrong} & None & \multirow{4}{*}{63.0\,{\scriptsize$\pm$2.3}} & \cellcolor[RGB]{214,128,130}76.8\,{\scriptsize$\pm$2.7} & \multirow{4}{*}{55.2\,{\scriptsize$\pm$3.8}} & \cellcolor[RGB]{153,179,205}45.8\,{\scriptsize$\pm$5.2} \\
 & Collab. &  & \cellcolor[RGB]{214,128,130}82.8\,{\scriptsize$\pm$1.7} &  & \cellcolor[RGB]{240,209,203}\textbf{58.9}\,{\scriptsize$\pm$4.7} \\
 & Conform. &  & \cellcolor[RGB]{214,128,130}\textbf{87.2}\,{\scriptsize$\pm$2.1} &  & \cellcolor[RGB]{125,160,196}31.3\,{\scriptsize$\pm$5.6} \\
 & Compet. &  & \cellcolor[RGB]{214,128,130}83.8\,{\scriptsize$\pm$1.0} &  & \cellcolor[RGB]{125,160,196}33.0\,{\scriptsize$\pm$2.9} \\
\bottomrule
\end{tabular}%
}
\caption{Different model family on LoCoBench (avg@3, \%, mean\,$\pm$\,SD over three rollouts), SkillOpt only. Mistral: weak = Ministral-3-8B, strong = Mistral-Small-4-119B-A6B (NVFP4); teams that use only the strong model serve it on two GPUs. Colors show the change from \emph{Pre} of the same team as in Table~\ref{tab:main}; \textbf{bold} marks the best structure of interdependence.}
\vspace{-5mm}
\label{tab:mistral}
\end{table}
In Table~\ref{tab:family}, both Gemma-4 agents fall in the 20--50 band on GAIA (27.5 and 44.2 alone), so the weakstrong team is a P2-like mix rather than the P3 mix of the Qwen pair, and Appendix~\ref{app:patterns} predicts small effects of evolution and structure. This holds across the teams. In weakstrong, evolution alone adds 1.7 points (37.5) and no structure helps further (30.8, 25.8, and 32.5 under collaboration, conformity, and competition); competition moves the chair after the first iteration, but to the weak answer\_agent rather than the strong web\_agent, and the team ends below none, as in OfficeQA Pro weakstrong (Section~\ref{sec:who}). The onlystrong team already outscores the strong agent alone before evolution (58.3 vs 44.2), but evolution barely changes it under any structure (50.0--59.2). In onlyweak, every structure stays within 1.7 points of \emph{Pre} (20.0) except competition (30.0), which hands the chair to web\_agent, the role that does most of the work; in the main GAIA onlyweak team, competition is likewise the best structure with SkillOpt and Meta-Team (28.3 and 27.5). Evolving either agent alone does not help either (28.3 and 43.3).

On LoCoBench, the Mistral pair (58.2 and 77.4 alone) falls in the same bands as the Qwen pair (Table~\ref{tab:mistral}). Two results carry over: collaboration is again the best or tied-best structure for the weakstrong and onlystrong teams (83.8 and 58.9), and competition again collapses the onlyweak team (45.5 vs.\ 71.4 under none, cf.\ 28.3 with Qwen). Unlike with Qwen, competition also lowers the weakstrong team (61.6 vs.\ 78.5 before evolution), and no structure consolidates the onlystrong team. Instead of moving the work to one role as in P5, the team grows from six to eight to eleven roles, and the share of final sessions that time out rises from 30\% before evolution to 43--53\% under none, conformity, and competition (45.8, 31.3, and 33.0). Only collaboration keeps timeouts at the Pre level (29\%) and improves on Pre. As in Section~\ref{sec:coordination}, added roles pay off only when they remove a bottleneck; here they mostly add waiting.

\section{Prompt Examples}\label{app:prompts}
We show the prompts that define the task, the team, and each structure of interdependence. All prompts are copied verbatim from the experiment code; values filled in by the code appear as placeholders in braces (e.g., \textit{\{partner\_name\}}), and the first line of each box gives the source file and line. Role prompts and constitutions are the weakstrong templates; across team compositions, only the \textit{model:} line of each role differs, except that the OfficeQA weakstrong constitution adds a \textit{Model split} paragraph. The prompts of SkillOpt and MemRL are used without modification from their public implementations and are omitted.

\subsection{Task Prompts}\label{app:prompt_task}
GAIA and OfficeQA require a \textit{FINAL ANSWER:} line, while LoCoBench is scored on the files written to \textit{solution/}. OfficeQA answers are judged by \textit{openai/gpt-5-mini} with the official HLE grading prompt.
\promptfile{Task prompts (GAIA, LoCoBench)}{code_task_prompts}
\promptfile{OfficeQA task prompt and LLM judge prompt}{code_officeqa_prompts}

\subsection{Constitutions and Chairman Prompts}\label{app:prompt_roles}
The chairman is plan\_agent in GAIA and planner in LoCoBench and OfficeQA. Prompts of the other roles are in the code repository.
\promptfile{GAIA constitution}{gaia_constitution}
\promptfile{GAIA plan\_agent (chairman)}{gaia_role_plan_agent}
\promptfile{LoCoBench constitution}{loco_constitution}
\promptfile{LoCoBench planner (chairman)}{loco_role_planner}
\promptfile{OfficeQA constitution (weakstrong)}{officeqa_constitution}
\promptfile{OfficeQA planner (chairman)}{officeqa_role_planner}

\subsection{Structure of Interdependence}\label{app:prompt_structures}
Each structure of interdependence differs in how agents discuss L2 notes, who decides on L3 changes, and, under competition, who leads.

\paragraph{L2 discussion.} Under none, L2 is skipped.
\promptfile{L2 discussion protocol: collaboration}{tmpl_l2_protocol_collaboration}
\promptfile{L2 discussion protocol: conformity}{tmpl_l2_protocol_conformity}
\promptfile{L2 discussion protocol: competition}{tmpl_l2_protocol_competition}

\paragraph{L3 proposals and decisions.} The chairman receives the structure prompt below. At the end of each iteration, the actual model of the chairman decides which proposals to keep in a single call (temperature 0); under none, all proposals are kept without a call.
\promptfile{L3 structure prompt (sent to the chairman)}{tmpl_l3_structure}
\promptfile{Decision: collaboration}{code_l3_decision_collaboration}
\promptfile{Decision: conformity}{code_l3_decision_conformity}
\promptfile{Decision: competition}{code_l3_decision_competition}

\paragraph{Chairman change under competition.} When another role becomes the chairman, each agent receives a notice; only the authority to lead passes to the new chairman.
\promptfile{Chairman change notice}{code_chairman_change_notice}

\subsection{Meta-Team Self-Reflection}\label{app:prompt_native_l1}
Under Meta-Team, each agent edits its own prompt patches and skills after a session with the prompt below.
\promptfile{L1 self-reflection prompt}{tmpl_l1_agent}

\section{Usage of AI Assistants}
All content of this paper was written by the authors. We used an AI assistant (ChatGPT) only to polish the language, such as improving clarity, correcting grammar, and enhancing readability. It was not involved in developing research ideas, designing experiments, interpreting results, or generating any content.

\end{document}